\documentclass{article}

\usepackage[preprint]{neurips_2026}
\usepackage[utf8]{inputenc} % allow utf-8 input
\usepackage[T1]{fontenc}    % use 8-bit T1 fonts
\usepackage{hyperref}       % hyperlinks
\usepackage{url}            % simple URL typesetting
\usepackage{booktabs}       % professional-quality tables
\usepackage{amsfonts}       % blackboard math symbols
\usepackage{nicefrac}       % compact symbols for 1/2, etc.
\usepackage{microtype}      % microtypography
\usepackage{xcolor}         % colors
\usepackage{graphicx}
\usepackage{amsmath}
\usepackage{amssymb}
\usepackage{makecell}
\usepackage{threeparttable}
\usepackage{multirow}
\usepackage{caption}
\usepackage{float}      
\usepackage{tabularx}   
\usepackage{pdflscape}
\usepackage{booktabs}
\usepackage{algorithm}
\usepackage{algpseudocode}
\usepackage{placeins}

\title{PALIN: Pixel-Aligned LiDAR--NIR Multimodal 3D Object Detection with Benchmark}

\author{%
  \normalfont
  Yanan Zhang \quad
  Amit Vishnia \quad
  Chengbo Ai \quad
  Eleni Christofa \\
  Runan Cheng \quad
  Siyuan Meng \quad
  Yuanchang Xie \quad
  Yu-Min (Thomas) Yang \\[0.8em]
  {\small University of Massachusetts Amherst
  \quad TrAnalytics LLC}\\[0.4em]
  {\small\texttt{yananzhang@umass.edu}}
}

\begin{document}

\maketitle

\begin{abstract}
3D object detection is critical for intelligent transportation systems. LiDAR has been widely adopted for accurate geometric measurements; however, purely geometric representations remain ambiguous under sparse observations and occlusions, while cross-sensor multimodal fusion suffers from calibration and synchronization errors. To overcome these challenges, we leverage an Ouster LiDAR to construct PALIN (Pixel-Aligned LiDAR–NIR), a large-scale intrinsically pixel-aligned multimodal dataset jointly capturing geometry and NIR intensity, comprising approximately 8,000 annotated frames from static urban intersections and diverse mobile suburban scenes. Building upon this dataset, we propose CMambaFusion, a novel fusion framework consisting of a four-directional Cross-Mamba module for bidirectional cross-modal interaction and a Gated Concat-Mamba module for joint feature refinement. Extensive experiments demonstrate that PALIN enables effective multimodal learning, and CMambaFusion achieves superior accuracy–efficiency trade-offs over both LiDAR-only and representative fusion baselines.
\end{abstract}
% ========================================================
\section{Introduction}
3D object detection is a fundamental capability for intelligent transportation systems (ITS), enabling traffic monitoring, safety analysis, and autonomous driving perception~\cite{review}. LiDAR sensors have been widely adopted for their accurate depth measurements~\cite{survey}, yet purely LiDAR-based perception suffers from point sparsity, occlusions, and the absence of appearance cues~\cite{deep}. Although multimodal fusion combining LiDAR with image-based modalities alleviates these limitations, existing approaches are typically built on heterogeneous sensor configurations requiring accurate extrinsic calibration and temporal synchronization, introducing spatial inconsistencies that limit fusion effectiveness~\cite{deep}.

Recent advances in LiDAR sensor technology offer a promising alternative. Certain LiDAR devices (e.g., Ouster  \footnote{Ouster, Inc., \url{https://ouster.com}}) simultaneously output multiple co-registered measurement channels from a single laser firing, including range, reflectivity, signal, and Near-Infrared (NIR) intensity~\cite{ouster_docs}. Because all channels are produced within the same sensing process, geometric and NIR observations are intrinsically aligned at the pixel level, eliminating cross-sensor calibration and synchronization errors. Among these channels, NIR intensity provides appearance, texture, and material cues complementary to geometric measurements.

Building upon this capability, we construct PALIN, a large-scale intrinsically aligned LiDAR--NIR multimodal dataset containing 8,587 annotated frames from both mobile and roadside sensing platforms, covering diverse traffic scenarios including urban intersections, suburban roads, and campus environments. The native pixel-level correspondence between modalities enables systematic study of intrinsic multimodal 3D detection without cross-sensor alignment ambiguity.

Nevertheless, constructing an effective fusion architecture for this modality pair remains non-trivial. Convolutional models are limited by local receptive fields, while Transformer-based methods suffer from quadratic complexity over high-resolution feature maps. State-space models (SSMs), particularly Mamba, achieve global receptive fields with linear complexity~\cite{sigma}, but extending Mamba to structured multimodal fusion requires addressing two challenges: enabling bidirectional cross-modal information exchange and consolidating enhanced features into a compact unified representation.  A single-stage design cannot simultaneously satisfy both requirements, cross-modal interaction without subsequent joint refinement leaves complementary information fragmented across separate feature streams, while direct concatenation without prior cross modal conditioning fails to establish meaningful inter-modality correspondence.

To this end, we propose CMambaFusion, a two-stage architecture consisting of a four-directional Cross-Mamba module for cross-modal interaction and a Gated Concat-Mamba module for compact joint feature refinement, enabling effective multimodal fusion with linear computational complexity.

Our main contributions are as follows:

\begin{itemize}
\item \textbf{PALIN Dataset.}
We introduce PALIN, a large-scale pixel-aligned LiDAR--NIR dataset with 8,587 annotated frames providing native pixel-level correspondence across both mobile and roadside configurations, enabling intrinsic multimodal 3D detection without calibration.

\item \textbf{CMambaFusion.}
We propose CMambaFusion, a two-stage Mamba-based fusion module comprising a four-directional Cross-Mamba for cross-modal interaction and a Gated Concat-Mamba for compact joint refinement, with linear-complexity global cross-modal modeling.

\item \textbf{Benchmark and Evaluation.}
We establish the first multimodal detection benchmark on PALIN, integrating CMambaFusion into a complete detection pipeline and comparing against LiDAR-only and fusion methods. CMambaFusion achieves superior accuracy-efficiency trade-offs, validating the effectiveness of both the dataset and the fusion design.
\end{itemize}

% ========================================================
\section{Related Work}
\subsection{3D Perception Datasets}
Existing outdoor 3D object detection datasets can be broadly categorized by their sensor configurations. Early benchmarks such as KITTI~\cite{kitti} pair a single LiDAR with stereo RGB cameras, while large-scale datasets including Waymo~\cite{waymo} and PandaSet~\cite{pandaset} extend this paradigm with multiple synchronized cameras and LiDAR units, enabling richer scene coverage at the cost of increased cross-sensor alignment complexity. Beyond LiDAR--camera setups, nuScenes~\cite{nuscenes} provides a comprehensive sensor suite with cameras, LiDAR, and radars, while View-of-Delft~\cite{vod} combines LiDAR, stereo cameras, and radar for dense urban detection. These datasets all rely on heterogeneous sensor configurations where cross-modal alignment is achieved through calibration and reprojection.

In contrast, PALIN is constructed from an Ouster OS1-128 that simultaneously outputs intrinsically pixel-aligned geometric and NIR measurements from the same laser firing, eliminating cross-sensor calibration while providing privacy-preserving appearance cues suitable for roadside deployment. To our knowledge, PALIN is the first large-scale outdoor 3D detection benchmark with native pixel-level LiDAR--NIR correspondence across both mobile and roadside sensing configurations.

\subsection{LiDAR-based 3D Object Detection}
LiDAR-based 3D detectors can be broadly categorized by their point cloud representation strategy. Point-based methods such as PointNet++ operate directly on raw point sets and preserve fine-grained geometric detail, but scale poorly to large outdoor scenes. Grid-based methods address this limitation by converting point clouds into structured representations. VoxelNet~\cite{voxelnet} pioneers end-to-end voxel feature learning, while SECOND~\cite{second} introduces sparse convolutions to accelerate voxel-based processing. PointPillars~\cite{pointpillars} further simplifies the pipeline by encoding points into vertical pillars and producing a BEV pseudo-image, achieving a favorable speed--accuracy trade-off well-suited for real-time deployment. CenterPoint~\cite{centerpoint} extends BEV-based detection with a keypoint-driven head that elegantly handles orientation estimation. VoxelNeXt~\cite{voxelnext} pushes the efficiency frontier with a fully sparse voxel architecture that eliminates the need for dense BEV feature maps. Despite these advances, purely geometric representations remain vulnerable to point sparsity at long range, occlusions, and lack of appearance cues, motivating the use of complementary sensing modalities.

\subsection{Multimodal Fusion for 3D Detection}
LiDAR-camera fusion methods can be organized by the stage at which cross-modal information is introduced. Input-level methods such as PointPainting~\cite{pointpainting} and PointAugmenting~\cite{pointaugmenting} enrich LiDAR points with projected image features prior to 3D encoding. Feature-level methods such as DeepFusion~\cite{deepfusion} aggregate image features into LiDAR representations via cross-attention, while TransFusion~\cite{transfusion} generates LiDAR-derived queries refined over image feature maps. BEV-centric frameworks such as BEVFusion~\cite{bevfusion} fuse lifted camera features with LiDAR BEV representations, and bidirectional architectures including BAFusion~\cite{bafusion} and BiCo-Fusion~\cite{bicofusion}. LIFT~\cite{lift} enable mutual cross-modal attention with temporal extensions. Despite their diversity, these approaches rely on heterogeneous sensors and require accurate extrinsic calibration and geometric reprojection, introducing alignment uncertainty.

\subsection{Mamba-based Fusion for Multimodal Perception}
State space models (SSMs), particularly Mamba~\cite{mamba}, provide efficient sequence modeling with linear computational complexity via context-aware selective state updates. SSMs have been extended to visual tasks through bidirectional and 2D scanning mechanisms~\cite{visionmamba, vmamba}, and applied to medical image segmentation, image restoration, and point cloud analysis~\cite{umamba, mambair, pointmamba}, achieving competitive performance at substantially lower complexity than Transformers.

For multimodal fusion, RemoteDet-Mamba~\cite{remotedet} applies directional selective scanning for lightweight RGB-thermal detection. DDHFusion~\cite{ddhfusion} introduces cross-modal Mamba modules in the BEV domain, though SSM parameters are jointly derived from concatenated features and state transitions remain modality-independent, limiting the depth of cross-modal feature exchange. MambaFusion~\cite{mambafusion} integrates reliability-aware fusion and token alignment into a state-space framework. Sigma~\cite{sigma} enables cross-modal interaction by exchanging output matrices $C$ between modalities during state-space decoding, but its design lacks structured multi-directional spatial propagation. In contrast, CMambaFusion explicitly incorporates four-directional scanning within the cross-modal interaction stage and consolidates the enhanced features into a compact representation via a gated mechanism.

\section{PALIN Dataset}
\subsection{Sensor Setup and Data Acquisition}
All data were collected using two Ouster OS1-128 LiDAR sensors mounted side by side on a shared platform. Each sensor provides a full $360^\circ$ horizontal field of view and a $42.4^\circ$ vertical field of view, outputting co-registered channels (range, reflectivity, signal, and NIR) with intrinsic pixel-level alignment. The dataset contains 8,587 annotated daytime frames from two complementary sensing configurations (see Appendix~\ref{app:dataset} and Table~\ref{tab:app_sequences} for detailed acquisition parameters).
 
\textbf{Bus-mounted mobile sensing.} Two sensors were mounted at the front-left and front-right of a transit bus at approximately 0.5\,m above ground, covering campus roads, suburban streets, and signalized intersections across western Massachusetts. After removing stationary and low-quality frames, 6,585 annotated frames across five bus routes are retained.
 
\textbf{Roadside static sensing.} Two sensors were deployed at approximately 2.0\,m height at a busy urban intersection in Boston, capturing dense traffic interactions from a fixed viewpoint across 2,002 annotated frames. Fig.~\ref{fig:app_dataset_overview} shows both sensing platforms and representative scenes.

\subsection{Intrinsic Pixel-Level Correspondence}
The OS1-128 produces a structured range-view of resolution $H \times W = 128 \times 1024$, where each pixel $(u, v)$ maps to a unique laser beam with a fixed azimuth and sensor-calibrated elevation angle. All measurement channels share this unified projection grid. Given a 3D point $\mathbf{p} = (x, y, z)$, the forward spherical projection is:
\begin{equation}
\phi = \arctan\!\left(\tfrac{y}{x}\right), \quad
\theta = \arcsin\!\left(\tfrac{z}{\|\mathbf{p}\|}\right)
\end{equation}
\begin{equation}
u = \mathrm{round}\!\left(\tfrac{\phi}{2\pi} \cdot W\right), \quad
v = \arg\min_{k} \left| \theta - \alpha_k \right|
\end{equation}
where $\phi \in [-\pi, \pi)$ is the azimuth angle, $\theta$ the elevation angle, and $\{\alpha_k\}_{k=0}^{H-1}$ the calibrated beam elevation angles. Because all channels share this projection, each NIR pixel $(u,v)$ corresponds directly to its associated 3D point, eliminating cross-sensor calibration and synchronization errors. The inverse mapping is provided in Appendix~\ref{app:dataset}. As a further consequence of this correspondence, projected 2D bounding boxes in the NIR image can be derived directly from the 3D annotated cuboids without any additional calibration step; representative examples are shown in Fig.~\ref{fig:app_modality_complementarity}.

\subsection{Annotation, Ontology, and Statistics}
All frames are annotated with oriented 3D bounding boxes defined in the coordinate frame of the left-mounted sensor, using the Segments.ai platform\footnote{Segments.ai: \url{https://segments.ai}}, covering 11 object categories: \textit{car, bus, medium-size truck, pickup truck, semi-truck, person, cyclist, scooter, motorcycle, train,} and \textit{towed object}. Each instance is labeled with category, 3D position, dimensions, yaw orientation, and auxiliary attributes including occlusion level, difficulty (Easy/Moderate/Hard), and modality dominance—indicating whether an object is primarily observable from LiDAR geometry, NIR appearance, or both. The dataset exhibits a long-tailed distribution with passenger vehicles dominating; Fig.~\ref{fig:app_dataset_statistics} summarizes the category distribution, per-class frame count, and object length distribution. Full ontology definitions and per-class illustrations are provided in Appendix~\ref{app:dataset}.

\subsection{Comparison with Existing Multimodal Datasets}
Table~\ref{tab:app_dataset_comparison} compares PALIN with representative public 3D perception datasets. Unlike conventional LiDAR--camera datasets, PALIN offers four key advantages. First, its intrinsic single-sensor multimodality provides native pixel-level correspondence without calibration, synchronization, or reprojection errors. Second, the NIR modality supplies complementary appearance and texture cues that geometry alone cannot capture, with modality dominance varying by scene and object type. Third, the novel modality-dominance attribute explicitly annotates whether each object is primarily observable from geometry, NIR, or both, enabling fine-grained evaluation of fusion methods. Finally, the combination of mobile bus-mounted and static roadside sensing provides both broad spatial coverage and dense urban traffic interaction scenarios—making PALIN uniquely suited for studying intrinsic multimodal perception under controlled alignment conditions.

\section{Method}

% ── 1. Overview ───────────────────────────────────────────────────────
\subsection{Overview}
\begin{figure*}[t]
    \centering
    \includegraphics[width=\textwidth]{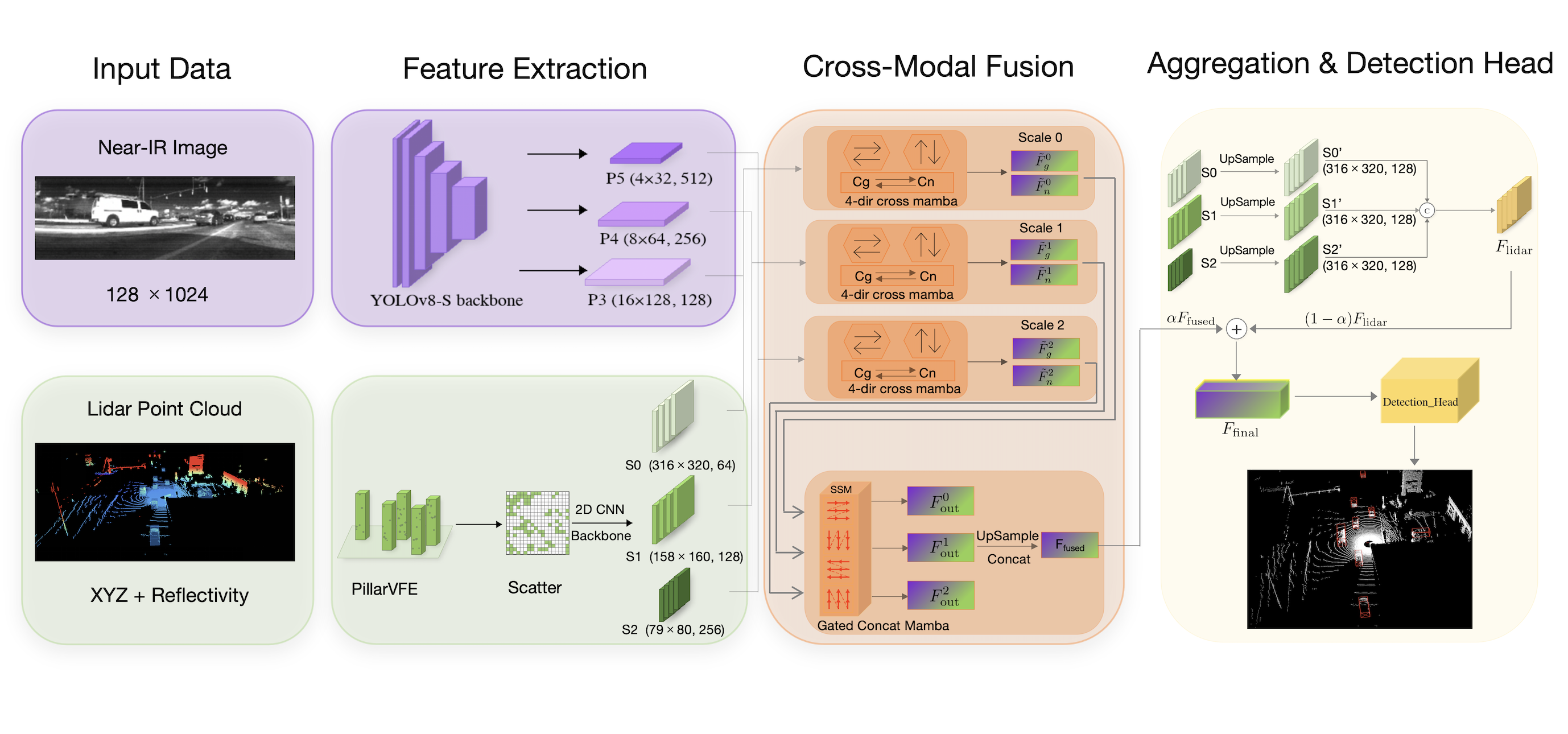}
     \caption{Overall framework of PALIN multimodal 3D detection. }
    \label{fig:framework}
\end{figure*}

Given a LiDAR point cloud and an intrinsically aligned NIR image, our goal is to perform multimodal 3D object detection (Fig.~\ref{fig:framework}). We extract modality-specific features using two dedicated backbones. For the LiDAR modality, we adopt PointPillars~\cite{pointpillars} to produce multi-scale geometric features $\{\mathbf{F}_g^{s}\}_{s=0}^{2}$, which efficiently encodes sparse 3D points into a structured BEV representation that preserves horizontal spatial layout for spatial modeling and feature-level fusion. For the NIR modality, we employ a YOLOv8-S~\cite{yolov8} backbone pretrained on PALIN NIR images to produce multi-scale feature maps $\{ \mathbf{F}_n^{s} \}_{s=0}^{2}$, leveraging its effectiveness for NIR object detection~\cite{singh2025nir}, and its computational efficiency for multi-scale extraction. 

 To fuse the two modalities, we introduce CMambaFusion, which is applied independently at three feature scales. Multi-scale fusion is adopted to capture complementary information at different granularities, fine scales encode local details such as edges and textures, while coarse scales capture higher-level semantic context. At each scale $s$, the feature pair $(F_g^s, F_n^s)$, is first processed by the 4-Dir Cross-Mamba module to perform cross-modal interaction, producing cross-modally enhanced features $\tilde{\mathbf{F}}_g^{s}$, $\tilde{\mathbf{F}}_n^{s}$, and then refined by the Gated Concat-Mamba module, producing a compact fused representation $F_{\mathrm{out}}^s$.
The per-scale output $F_{\mathrm{out}}^s$ is upsampled and concatenated to form $F_{\mathrm{out}}$. This fused feature is further combined with the original geometric feature $F_{\mathrm{lidar}}$ via a learnable scalar $\alpha$, and the result is fed into an anchor-based detection head~\cite{ssd,pointpillars}.

% ── 2. Feature Extraction ─────────────────────────────────────────────
\subsection{Feature Extraction}

\noindent\textbf{Geometric Encoder.} The geometric encoder consists of a PillarVFE module, a pillar scatter, and a lightweight 2D CNN backbone following PointPillars~\cite{pointpillars}. Raw point cloud features $(x,y,z,\text{reflectivity})$ are encoded by the Pillar Feature Network and scattered into a BEV feature map over the detection range $[-51.2,51.2]\text{ m} \times [-60,41.12]\text{ m}$ with pillar size $0.16 \times 0.16 \text{ m}$. A 2D convolutional backbone with three stride-2 downsampling blocks, each comprising multiple $3\times3$ convolutions, produces multi-scale geometric features $\{F_g^{s}\}_{s=0}^{2}$ at spatial resolutions and channel dimensions $(316\times320, 64)$, $(158\times160, 128)$, and $(79\times80, 256)$, respectively.

\noindent\textbf{NIR Backbone.}
The NIR feature extractor consists of a CSPDarknet 
backbone~\cite{cspnet} and a PANet-style neck~\cite{panet}, following the YOLOv8-S~\cite{yolov8} architecture pretrained on the PALIN NIR images. For the $128\times1024$ input, multi-scale features are extracted from the PANet-style neck at feature strides $\{8,16,32\}$, yielding multi-scale NIR features $\{F_n^{s}\}_{s=0}^{2}$ with dimensions  $(16\times128, 128)$, $(8\times64, 256)$, and $(4\times32, 512)$, capturing fine-grained, intermediate, and high-level semantic features, respectively.

To facilitate cross-modal fusion, features from both modalities are aligned at each scale. Specifically, all features are first projected to a unified channel dimension $d=128$ via $1\times1$ convolutions. The geometric features are then downsampled to a common spatial resolution of $32\times64$ via adaptive average pooling, while the NIR features are resized through interpolation or pooling. The resulting aligned feature pairs $(F_g^{s}, F_n^{s}) \in \mathbb{R}^{32\times64\times128}$ are used as inputs to the subsequent fusion module. 

\subsection{Cross-Modal Fusion Modules}
CMambaFusion consists of two sequential stages, a 4-Dir Cross-Mamba module for bidirectional cross-modal interaction, followed by a Gated Concat-Mamba module for compact joint refinement. Both stages operate independently at each of the three feature scales; the scale index $s$ is omitted for brevity. Algorithm~\ref{alg:cmambafusion} summarizes both stages side by side.
 
\textbf{4-Dir Cross-Mamba} embeds four-directional 2D selective scanning directly within the cross-modal interaction step. Unlike Sigma~\cite{sigma}fusion approaches that rely on prebuilt global context from Mamba encoder backbones, our design accumulates spatial context at the fusion stage itself, enabling effective cross-modal interaction even when paired with purely convolutional backbones. Scanning along four directions ensures that spatially adjacent positions remain connected along at least one path, preventing cross-modal guidance from being attenuated over long sequential distances induced by spatial flattening. Given aligned feature maps $F_g, F_n \in \mathbb{R}^{H_f \times W_f \times d}$, each modality is projected and locally enhanced via depthwise convolution, then unfolded into four directional sequences. Cross-modal interaction is achieved by swapping the output matrix $C$ between modalities during SSM decoding for every direction, the geometric state is read out using the NIR context $C_n^k$, and vice versa. Directional outputs are merged back to 2D via inverse scan and directional summation, then residually added, yielding $\tilde{F}_g$ and $\tilde{F}_n$.

 \textbf{Gated Concat-Mamba} takes $\tilde{F}_g, \tilde{F}_n$ and produces a compact unified refinement representation. The modalities are concatenated and compressed by a $1{\times}1$ convolution to form a base representation $X_0$. In parallel, a separate $1{\times}1$ projection with expanded channel dimension $rd$ forms a unified single-stream representation $Z$, removing modality boundaries so the subsequent SSM can model long-range dependencies over a semantically coherent feature space. $Z$ passes through depthwise convolution and a four-directional single-stream SSM, with directional outputs merged via inverse scan and summation to produce a global refinement signal $Y$. A spatial gate $g$, derived from the cross-modal difference $\tilde{F}_g - \tilde{F}_n$ via a sigmoid-activated $1{\times}1$ convolution, amplifies refinement where inter-modal discrepancy is high. The output is $F_{\mathrm{out}} = X_0 + g \odot Y$.
 
\begin{figure}[t]
\centering
\includegraphics[width=\linewidth]{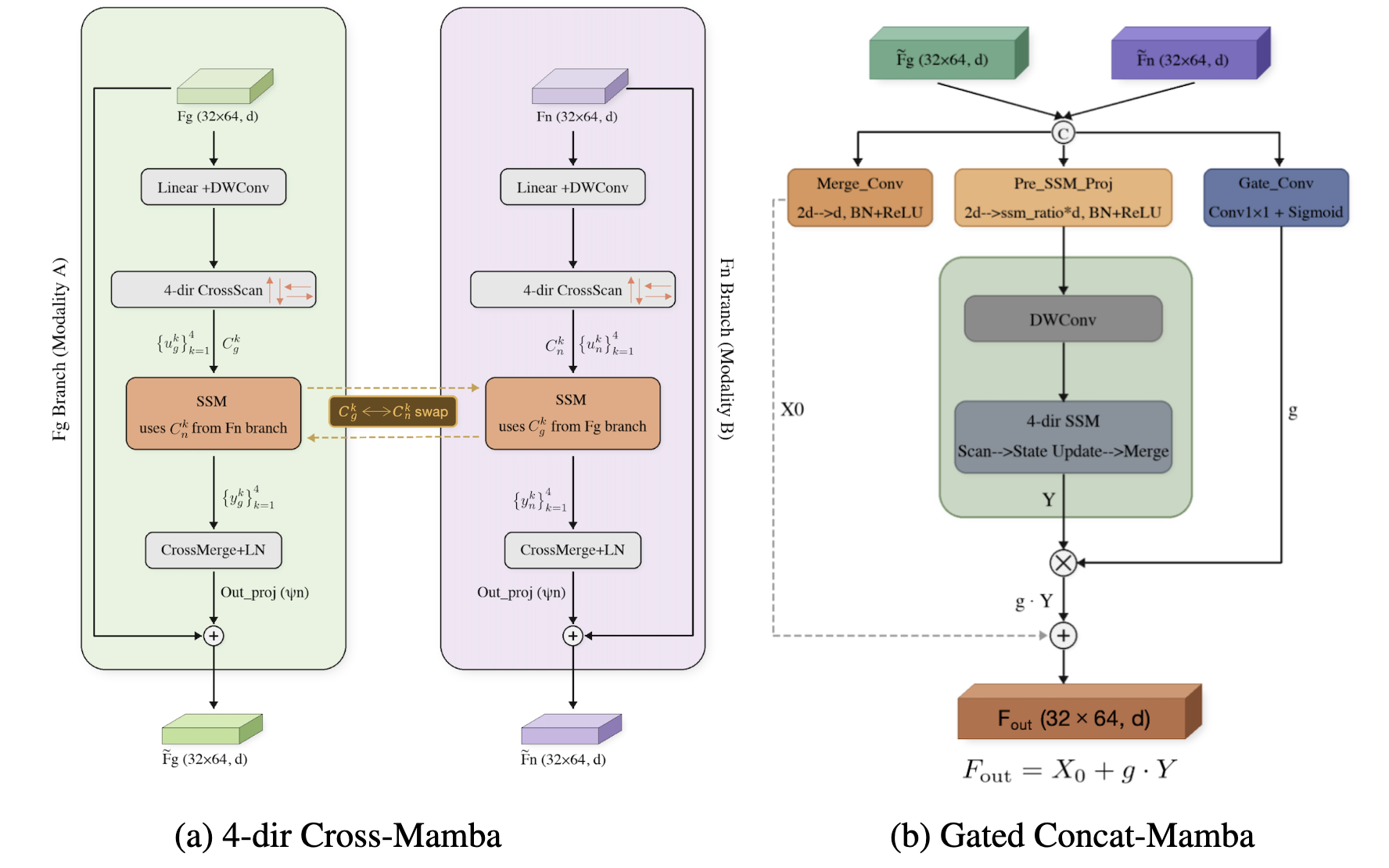}
\caption{Architecture of the proposed modules. (a) 4-dir Cross-Mamba: four-directional cross-modal state propagation. (b) Gated Concat-Mamba: feature refinement with difference-based gating.}
\label{fig:mamba_modules}
\end{figure}

% ── 4. Aggregation, Detection, and Training ───────────────────────────
\subsection{Aggregation and Detection Head}
The fused output at each scale $F_{\mathrm{out}}^{s}$ is upsampled to a shared spatial resolution and concatenated along the channel dimension to form the aggregated multimodal representation $F_{\mathrm{out}}$. The $F_{\mathrm{lidar}}$ is the concatenation of the three upsampled lidar BEV backbone outputs at uniform channel dimension 128. They are combined via a learnable scalar $\alpha$, which adaptively balances the multimodal and geometric features according to scene conditions. The resulting feature is fed into an anchor-based detection head~\cite{ssd,pointpillars} covering all 11 object categories. 

The detection loss follows PointPillars~\cite{pointpillars}, $\mathcal{L}_{\mathrm{det}} = \lambda_{\mathrm{cls}}\mathcal{L}_{\mathrm{cls}} + \lambda_{\mathrm{loc}}\mathcal{L}_{\mathrm{loc}} + \lambda_{\mathrm{dir}}\mathcal{L}_{\mathrm{dir}}$, where $\mathcal{L}_{\mathrm{cls}}$ is focal loss ~\cite{focal} for classification, $\mathcal{L}_{\mathrm{loc}}$ supervises 3D box regression ~\cite{fast}, and $\mathcal{L}_{\mathrm{dir}}$ is binary cross-entropy for orientation disambiguation. To promote semantic alignment of different modality features, enabling the model to learn cross-modal spatial correspondence without physical projection constraints, we add a cosine alignment loss over the enhanced feature pairs at all three scales: \begin{equation} \mathcal{L}_{\mathrm{align}} = 1 - \frac{1}{S}\sum_{s=0}^{S-1}\cos\!\left(\tilde{F}_g^{(s)},\,\tilde{F}_n^{(s)}\right), \qquad \mathcal{L} = \mathcal{L}_{\mathrm{det}} + \lambda_{\mathrm{align}}\mathcal{L}_{\mathrm{align}}. \end{equation}

\section{Experiments}
\subsection{Experimental Setup}

\noindent\textbf{Dataset and Split.}
All experiments are conducted on the proposed PALIN dataset.
The dataset is organized into two experimental settings.
The first setting uses only the mobile bus-mounted subset (dynamic-only), comprising 4{,}662 training frames, 583 validation frames, and 583 test frames.
The second setting incorporates both the bus-mounted and static Boston intersection subsets (combined), yielding 6{,}264 training frames, 783 validation frames,
and 783 test frames. The dataset contains 903 frames with no annotated objects, distributed across splits as background samples. All splits follow an 8:1:1 ratio, partitioned at the frame level.

\noindent\textbf{Evaluation Metrics.}
We adopt the KITTI AP$_{40}$ protocol~\cite{kitti,simonelli} with IoU thresholds of 0.7/0.5 for vehicle-type categories and 0.5/0.25 for small objects (Pedestrian, Cyclist, Scooter). Results are reported for 8 evaluable categories, full evaluation protocol details are provided in Appendix~\ref{app:impl}.

\noindent\textbf{Implementation Details.} 
All models are trained for 80 epochs with Adam (lr=$3 \times 10^{-3}$, weight decay=$10^{-2}$, batch size 4), with ground-truth sampling~\cite{second}, random horizontal flip, rotation in $[-\pi/4,\pi/4]$, and scaling in $[0.95, 1.05]$. All experiments use PyTorch 1.13.1 and OpenPCDet~\cite{openpcdet} on a single NVIDIA RTX 6000 Ada GPU (48\,GB). 

\subsection{Main Results}
Table~\ref{tab:main_results} reports the weighted mAP$_{40}$ of four methods under two training settings on PALIN.

\noindent\textbf{Full dataset setting (8k).}
The LiDAR-only PointPillars baseline achieves 71.20 3D@0.7 and 86.42 3D@0.5. Introducing bi-directional cross-attention fusion consistently improves performance, yielding gains of +0.56 in 3D@0.7 and +1.15 in 3D@0.5, confirming the complementary value of NIR features. The proposed 4-Dir Cross-Mamba further improves 3D@0.7 to 74.21 (+3.01 over baseline), demonstrating the advantage of structured four-directional cross-modal state propagation over attention-based fusion. Adding the Gated Concat-Mamba stage produces the best overall results with 74.97 3D@0.7 and 88.90 3D@0.5, achieving the highest scores in three out of four metrics.

\noindent\textbf{Dynamic-only setting (6k).}
Under the dynamic-only setting, the absolute performance is lower due to reduced training data and fewer static scene observations. However, the relative gain from CMambaFusion is more pronounced. The full model achieves +7.24 3D@0.7 and +5.44 3D@0.5 over the LiDAR-only baseline, compared to +3.01 and +2.48 in the full setting. This suggests that CMambaFusion is particularly effective when the LiDAR geometry alone is less discriminative, as is the case in mobile sensing scenarios with sparser point clouds and greater viewpoint variation.

% ================= Main Results Table (Fixed Position) =================
\begin{center}

\captionof{table}{Comparison of fusion methods on PALIN under two training settings (weighted mAP$_{40}$, \%).}
\label{tab:main_results}

\setlength{\tabcolsep}{4.5pt}
\renewcommand{\arraystretch}{1.15}

\resizebox{\linewidth}{!}{%
\begin{threeparttable}
\begin{tabular}{llcccc}
\toprule
Setting & Method & BEV@0.7 & 3D@0.7 & BEV@0.5 & 3D@0.5 \\
\midrule
\multirow{4}{*}{Full (8k)}
 & PointPillars (LiDAR-only) & 80.97 & 71.20 & 88.09 & 86.42 \\
 & + Cross-Attention         & 83.54 & 71.76 & 89.61 & 87.57 \\
 & + 4-Dir Cross-Mamba       & 84.56 & 74.21 & 90.41 & 88.30 \\
 & + CMambaFusion & \textbf{84.79} & \textbf{74.97} & \textbf{90.87} & \textbf{88.90} \\
\midrule
\multirow{4}{*}{Dynamic Only (6k)}
 & PointPillars (LiDAR-only) & 70.89 & 44.40 & 81.90 & 76.29 \\
 & + Cross-Attention         & 73.48 & 50.12 & 83.88 & 79.27 \\
 & + 4-Dir Cross-Mamba       & 74.81 & 51.51 & 84.07 & 81.20 \\
 & + CMambaFusion & \textbf{75.35} & \textbf{51.64} & \textbf{84.33} & \textbf{81.73} \\
\bottomrule
\end{tabular}

\begin{tablenotes}
  \footnotesize
  \item \textbf{Bold} denotes the best result per setting.
\end{tablenotes}

\end{threeparttable}
}

\end{center}

\noindent\textbf{Per-class analysis.}
Per-class results (Table~\ref{tab:perclass}) show consistent gains across multiple categories: Car (+5.25 at @0.7), Pedestrian (+1.17 at @0.7), and Scooter (+7.54 at @0.7), highlighting the benefit of NIR cues for small objects. Large vehicle categories (Pickup Truck, Medium Truck) benefit less, likely due to limited NIR discriminability when shapes are visually similar. Train shows a large gain (+23.07 at @0.7), reflecting strong geometric--appearance complementarity. 

\subsection{Comparison with SOTA}

As shown in Table~\ref{tab:sota}, SECOND and CenterPoint use the same geometric inputs for fair LiDAR-only comparison. For BEVFusion, we replace its Swin Transformer backbone with YOLOv8 for domain-appropriate NIR representations; its query-based head yields only 5.53\% at IoU=0.7 (34.06\% at IoU=0.5), while switching to an anchor-based head recovers performance to 71.50\%, indicating that the detection head is the primary bottleneck. We additionally implement a Bi-Cross-Attention Fusion baseline for fair attention-based comparison. CMambaFusion achieves 74.97\% 3D mAP$_{40}$, outperforming all compared methods.

\subsection{Ablation Studies}
Table~\ref{tab:ablation_all} ablates key design choices. All Cross-Mamba variants outperform the Cross-Attention baseline, confirming the effectiveness of directional SSM modeling. The number of scanning directions is the most critical factor: increasing from 1 to 4 yields the largest gain (+1.34), as multi-directional propagation is essential for spatial connectivity under sequence flattening. Adding the Concat-Mamba stage provides further improvement via unified single-stream refinement, while the sigmoid gate contributes a modest additional gain by adaptively modulating cross-modal differences. The full CMambaFusion achieves the best performance, confirming that all components are complementary.

\begin{center}

% left table
\begin{minipage}[t]{0.5\linewidth}
\centering
\captionof{table}{Comparison with SOTA methods on the PALIN dataset (3D AP$_{40}$ at IoU=0.7, \%).}
\label{tab:sota}
\setlength{\tabcolsep}{3pt}
\begin{tabular}{lc}
\toprule
Method & 3D mAP$_{40}$ \\
\midrule

\multicolumn{2}{c}{\textit{LiDAR-only}} \\
\midrule
PointPillars (baseline) & 71.20 \\
Second \cite{second} & 69.40 \\
CenterPoint \cite{centerpoint} & 71.43 \\

\midrule
\multicolumn{2}{c}{\textit{Fusion Methods}} \\
\midrule
BEVFusion~\cite{bevfusion}& 71.50 \\
Bi-Cross-Attention Fusion & 71.76 \\
Sigma~\cite{sigma} & 68.70 \\

\midrule
\textbf{CMambaFusion (Ours)} & \textbf{74.97} \\

\bottomrule
\end{tabular}
\end{minipage}
\hfill
% right table
\begin{minipage}[t]{0.48\linewidth}
\centering
\captionof{table}{Ablation study on scanning directions and module components (Full 8k, 3D mAP$_{40}$ at IoU=0.7, \%).}

\label{tab:ablation_all}
\setlength{\tabcolsep}{3pt}
\begin{tabular}{lccc}
\toprule
Method & Dir & GCM & @0.7 \\
\midrule
Cross-Attention  & -- & -- & 71.76 \\
Cross-Mamba & 1 & $\times$ & 72.87 \\
Cross-Mamba & 4 & $\times$ & 74.21 \\
CMambaFusion & 4 & $\checkmark$ & \textbf{74.97} \\
\bottomrule
\end{tabular}
\par\vspace{3pt}
\footnotesize Dir: number of scanning directions in Cross-Mamba;
\\GCM: Gated Concat-Mamba module.
\end{minipage}

\end{center}

\subsection{Efficiency Analysis}
Table~\ref{tab:efficiency} compares computational efficiency. The LiDAR-only baseline achieves 27.72 FPS with no fusion overhead. Compared to cross-attention, 4-Dir Cross-Mamba reduces fusion FLOPs from $\sim$1.4B to $\sim$12M ($>100\times$), highlighting the efficiency of linear $\mathcal{O}(L)$ SSMs over quadratic $\mathcal{O}(N^2)$ attention. At inference, 4-Dir Cross-Mamba achieves 9.6\% higher throughput and 9.55\% lower latency than cross-attention while delivering superior accuracy. Adding Gated Concat-Mamba incurs marginal overhead, maintaining a favorable accuracy–efficiency trade-off.

\begin{center}
\captionof{table}{
Efficiency comparison of fusion methods. 
}

\label{tab:efficiency}

\setlength{\tabcolsep}{4pt}

\begin{tabular}{lccccc}
\toprule
Method & Params & FPS & Latency & Mem & FLOPs \\
 & (M) & $\uparrow$ & (ms)$\downarrow$ & (GB)$\downarrow$ & (fusion) \\
\midrule
PointPillars        & 4.98  & 27.72 & 36.08 & 3.69 & -- \\
+ Cross-Attention   & 17.19 & 18.24 & 54.82 & 5.73 & $\sim$1.4B \\
+ 4-Dir Cross-Mamba & 17.34 & 19.99 & 50.04 & 5.74 & \textbf{$\sim$12M} \\
+ CMambaFusion      & 17.98 & 19.04 & 52.53 & 5.74 & \textbf{$\sim$12M} \\
\bottomrule
\end{tabular}

\end{center}

\subsection{Visualization}
Fig.~\ref{fig:pr_curve} shows PR curves for representative categories.  CMambaFusion achieves a more favorable precision–recall trade-off across multiple categories, for small objects (Pedestrian, Scooter), it maintains higher precision at medium-to-high recall, indicating that the integration of NIR appearance cues improves detection robustness without introducing excessive false positives. For large vehicles (Pickup Truck, Medium Truck) the improvement is less consistent, aligning with per-class quantitative results. This suggests that NIR features may provide limited discriminative information for these categories when object shapes are visually similar, especially under challenging conditions. 
 
Fig.~\ref{fig:app_qualitative} shows qualitative comparisons. In the bus-mounted scene, CMambaFusion recovers distant vehicles missed by the LiDAR-only baseline. In the Boston intersection scene, it produces additional true positives and reduces false detections, with more accurate localization. These results confirm that NIR appearance cues provide complementary contextual information that helps suppress incorrect predictions and recover missed objects.

\begin{center}

\includegraphics[width=\linewidth]{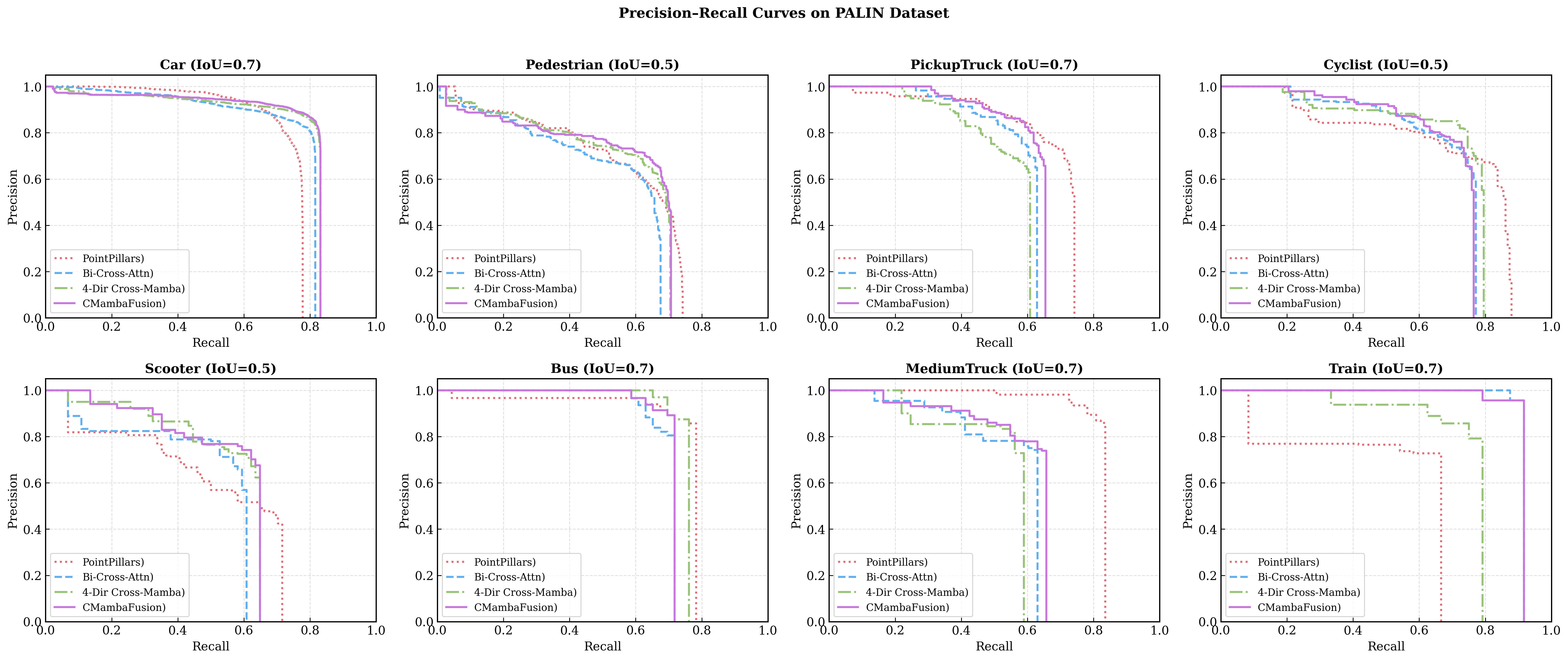}

\captionof{figure}{
Precision–recall (PR) curves for representative categories on the PALIN dataset.
}
\label{fig:pr_curve}

\end{center}

\section{Conclusion}
We present PALIN, a large-scale pixel-aligned LiDAR--NIR dataset, and CMambaFusion, a two-stage Mamba-based fusion framework combining four-directional Cross-Mamba for bidirectional cross-modal interaction and Gated Concat-Mamba for compact joint refinement. Extensive experiments show that CMambaFusion achieves superior accuracy--efficiency trade-offs over LiDAR-only and fusion baselines, with particularly pronounced gains in mobile sensing scenarios where LiDAR geometry alone is less discriminative.

Several limitations remain. The dataset is currently limited to daytime conditions, and only three of the four Ouster channels are utilized; the signal channel may offer additional complementary cues. Occlusion in dense urban scenes remains a key challenge, as severe multi-agent occlusion in the Boston subset degrades detection performance. CMambaFusion also shows less consistent gains for large vehicle categories where NIR discriminability is limited. Future work will explore occlusion-aware strategies, temporal modeling across frames, and stronger image backbones for large-vehicle recognition. We will publicly release the dataset and code to facilitate reproducible benchmarking and future research.

\bibliographystyle{unsrt}
\bibliography{references}

\clearpage
\appendix

% ================= A =================
\section{Dataset Details}
\label{app:dataset}

\renewcommand{\thefigure}{A\arabic{figure}}
\renewcommand{\thetable}{A\arabic{table}}
\setcounter{figure}{0}
\setcounter{table}{0}

% ---- A.1 Detailed Acquisition Parameters ----
\subsection{Detailed Acquisition Parameters}
\label{app:acquisition}

The dataset comprises six recording sequences in total, five bus-mounted routes (B43-01, B43-02, B43-03, 31-01, and 38-02) and one roadside deployment (Boston-01). Table~\ref{tab:app_sequences} summarizes the per-sequence recording conditions.

\textbf{Bus-mounted subset.} Data were recorded in January 2024 across Amherst and surrounding communities in western Massachusetts, including Hadley, North Amherst, Sunderland, South Deerfield, and Downtown Amherst, following three transit bus routes (Routes B43, 31, and 38). The sensor was mounted at approximately 0.5\,m above ground on the front-left of the bus. Data were captured at 10\,Hz and sparsely sampled at one frame per five, with stationary and low-quality frames removed to ensure scene diversity and annotation reliability.

\textbf{Boston roadside subset.} Data were recorded in October 2025 at a signalized intersection in Boston. The sensor was mounted at approximately 2.0\,m above ground on a fixed roadside structure. Data were captured at 20\,Hz and sampled at one frame per ten.

\begin{center}
\captionof{table}{Summary of PALIN recording sequences.}
\label{tab:app_sequences}
\setlength{\tabcolsep}{5pt}
\renewcommand{\arraystretch}{1.1}
\begin{tabular}{llccccc}
\toprule
Sequence & Platform & Site & Capture Hz & Sample Rate & Ann.\ Frames \\
\midrule
B43-01 & Bus & Amherst, Hadley & 10 & 1/5 & 768 \\
B43-02 & Bus & Hadley & 10 & 1/5 & 1027 \\
B43-03 & Bus & Amherst, Hadley & 10 & 1/5 & 1731 \\
31-01  & Bus & North Amherst, Sunderland, South Deerfield & 10 & 1/5 & 2302 \\
38-02  & Bus & Amherst  & 10 & 1/5 & 757 \\
\midrule
Boston-01 & Roadside & Commonwealth Ave @ Pleasant St, Brookline & 20 & 1/10 & 2,002 \\
\midrule
\multicolumn{5}{r}{\textbf{Total}} & \textbf{8,587} \\
\bottomrule
\end{tabular}
\end{center}

% ================= Fig. A1: Dataset Overview =================
\begin{center}

\begin{minipage}{0.48\linewidth}
    \centering
    \includegraphics[width=\linewidth]{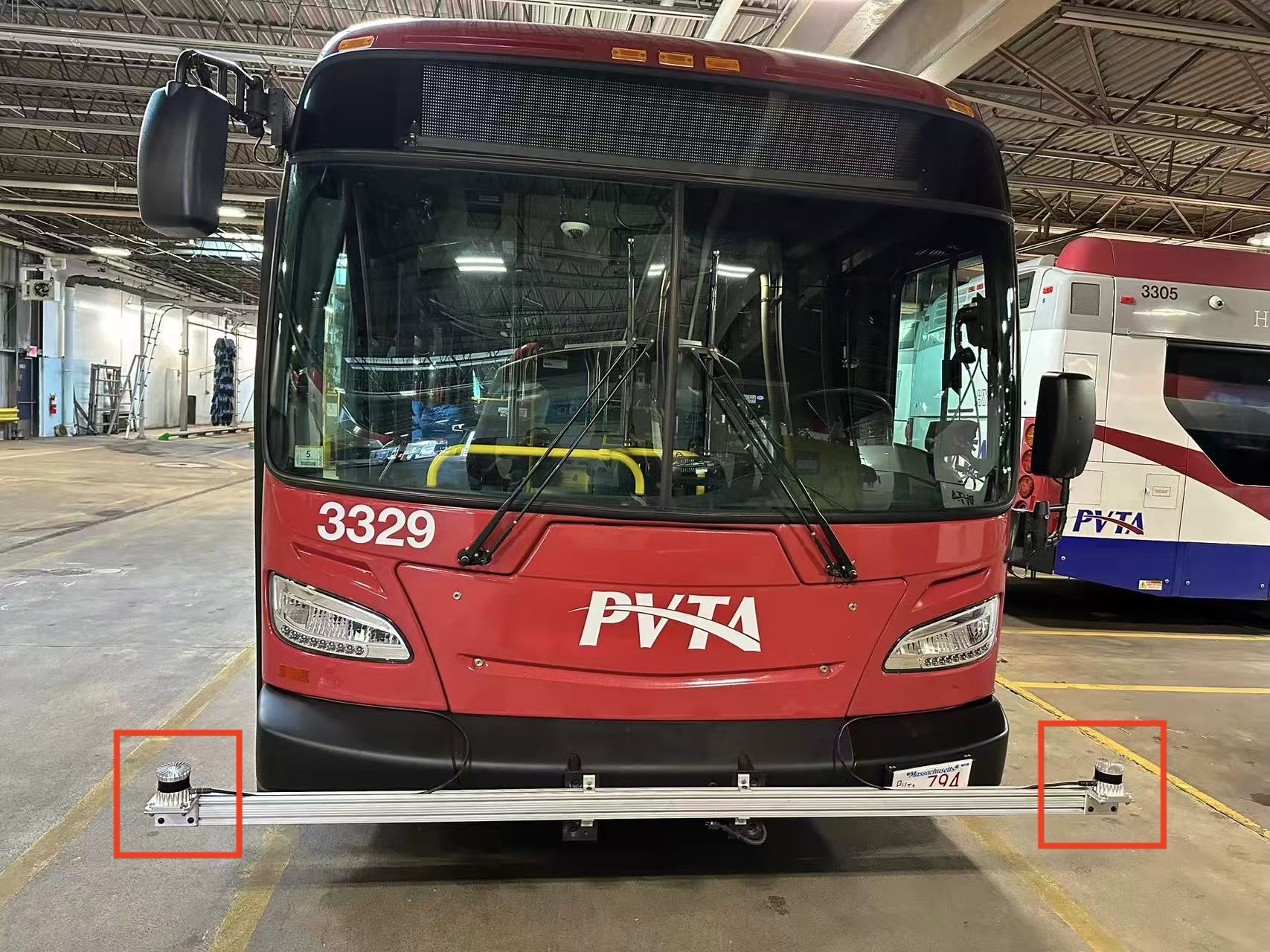}
    \\ (a) Bus-mounted mobile sensing platform.
\end{minipage}
\hfill
\begin{minipage}{0.48\linewidth}
    \centering
    \includegraphics[width=\linewidth]{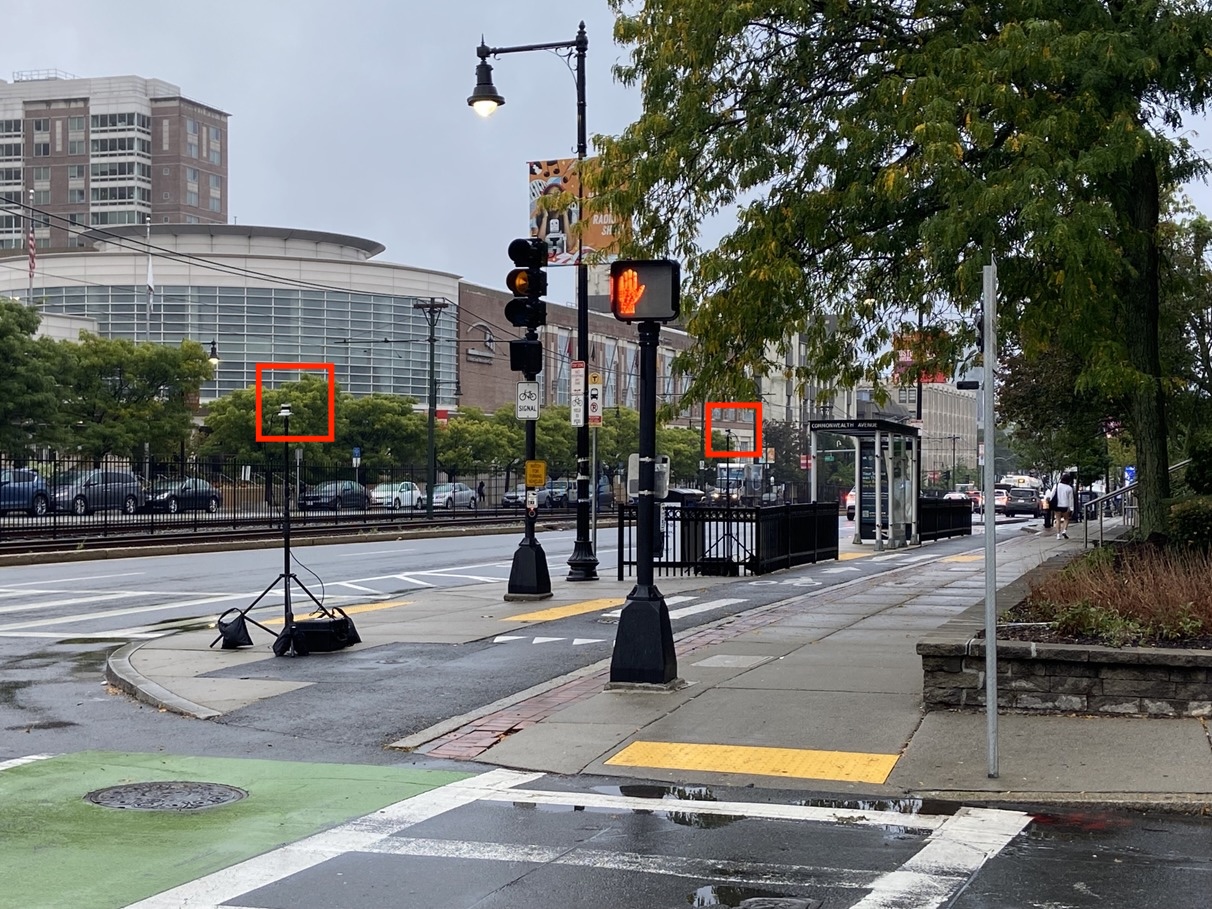}
    \\ (b) Static roadside sensing deployment.
\end{minipage}

\vspace{2mm}

\begin{minipage}{0.48\linewidth}
    \centering
    \includegraphics[width=\linewidth]{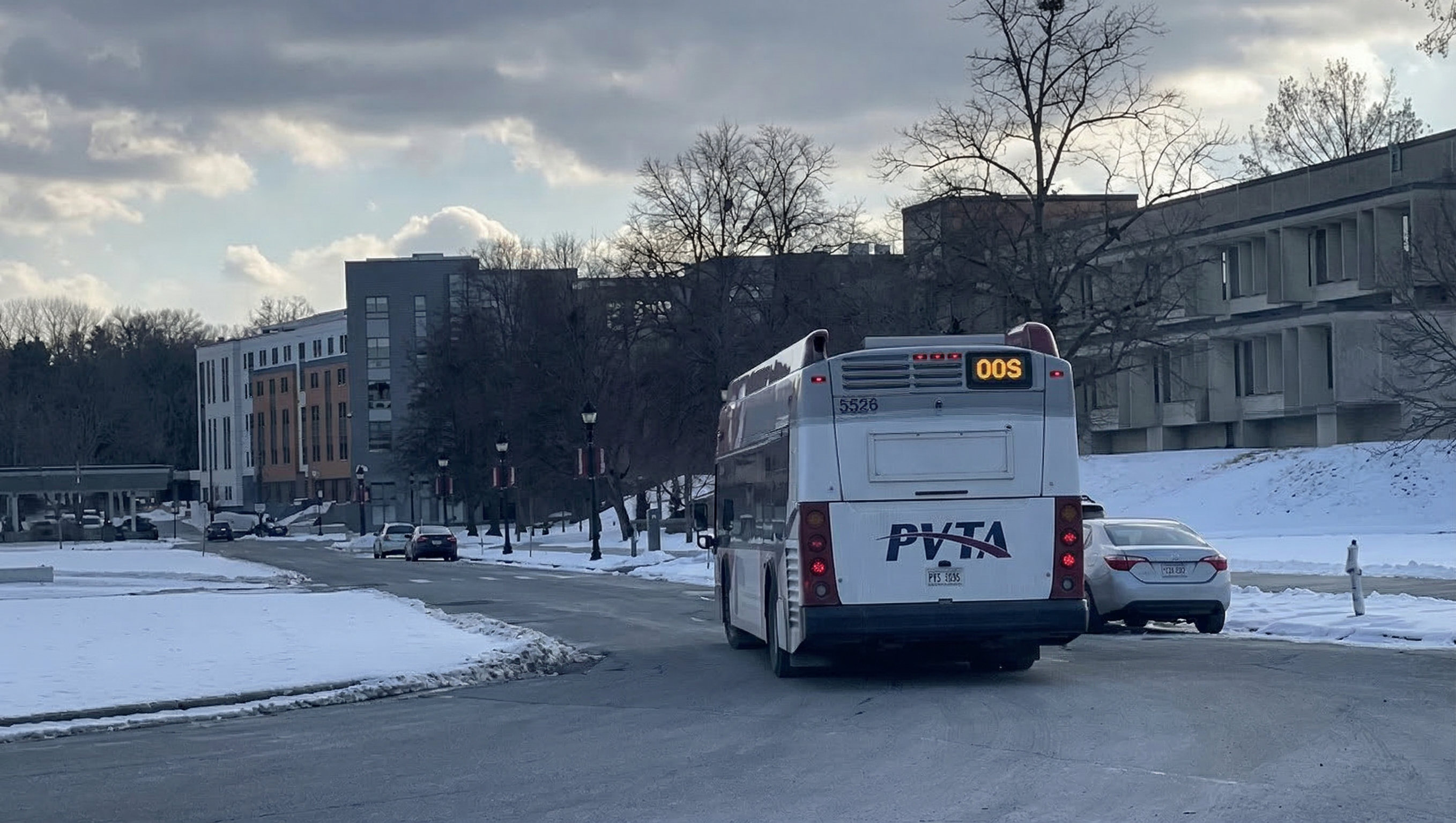}
    \\ (c) Example mobile traffic scene.
\end{minipage}
\hfill
\begin{minipage}{0.48\linewidth}
    \centering
    \includegraphics[width=\linewidth]{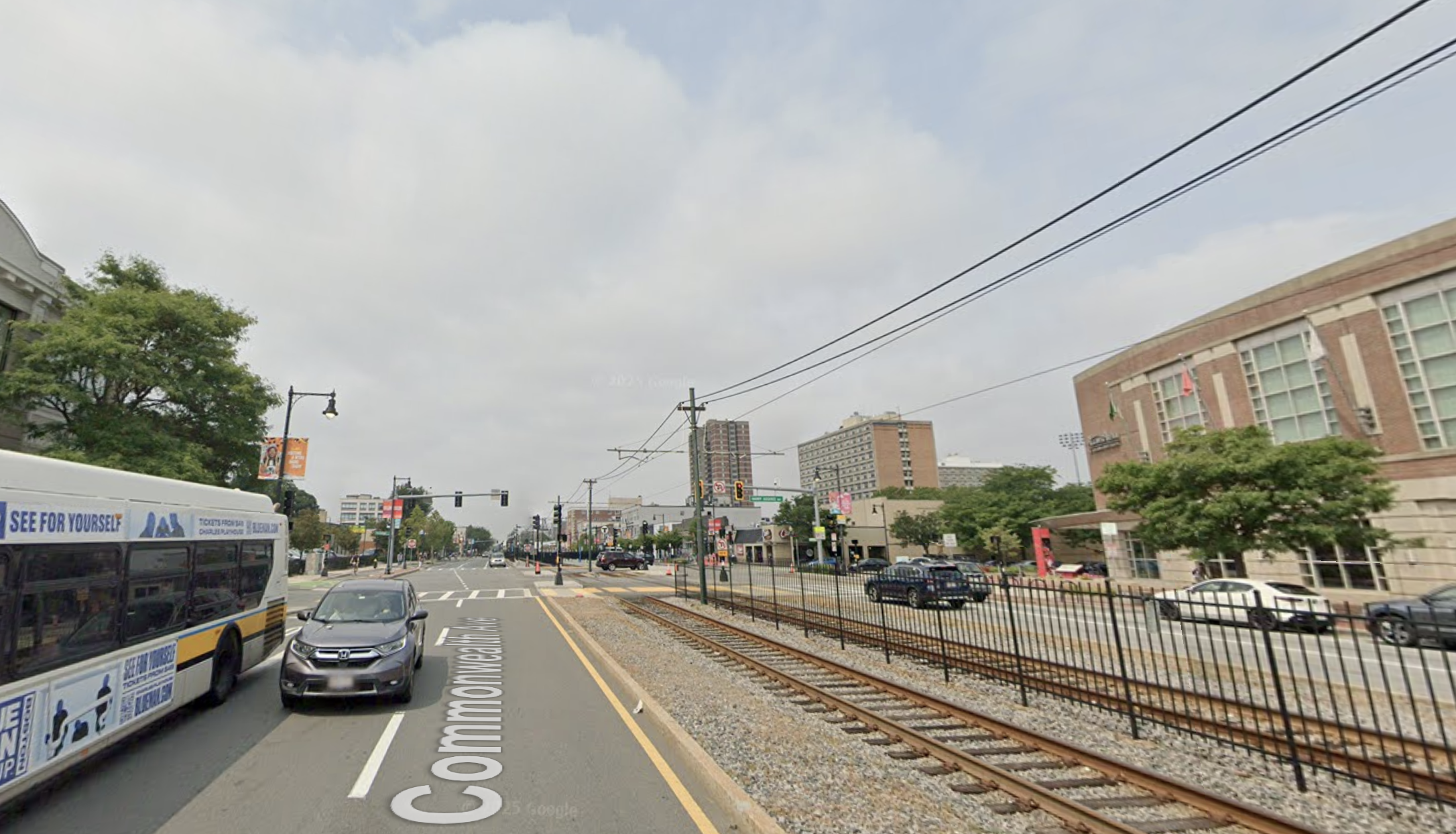}
    \\ (d) Example dense urban intersection scene.
\end{minipage}

\captionof{figure}{
Overview of the PALIN dataset, consisting of both bus-mounted mobile sensing
(Amherst subset) and static roadside deployment (Boston subset).
}
\label{fig:app_dataset_overview}

\end{center}

% ---- A.2 Inverse Pixel Mapping ----
\subsection{Inverse Pixel Mapping}
\label{app:inverse_mapping}

The forward spherical projection (Eqs.~1--2 in the main text) maps each 3D point to a pixel $(u, v)$ in the range-view image. The inverse mapping reconstructs the 3D point from a pixel coordinate and its associated range measurement $r$:
\begin{equation}
x = r \cos\theta \cos\phi, \quad
y = r \cos\theta \sin\phi, \quad
z = r \sin\theta,
\end{equation}
where $\phi$ is the azimuth angle recovered from $u$ via $\phi = (u / W) \cdot 2\pi$, and $\theta = \alpha_v$ is the calibrated beam elevation angle for row $v$. This inverse mapping is not used in our fusion pipeline, but is provided for completeness and to facilitate downstream research using the PALIN dataset.

% ================= Fig. A2: Modality Complementarity =================
\begin{center}

\begin{minipage}[t]{0.32\textwidth}
    \centering
    \includegraphics[width=\textwidth]{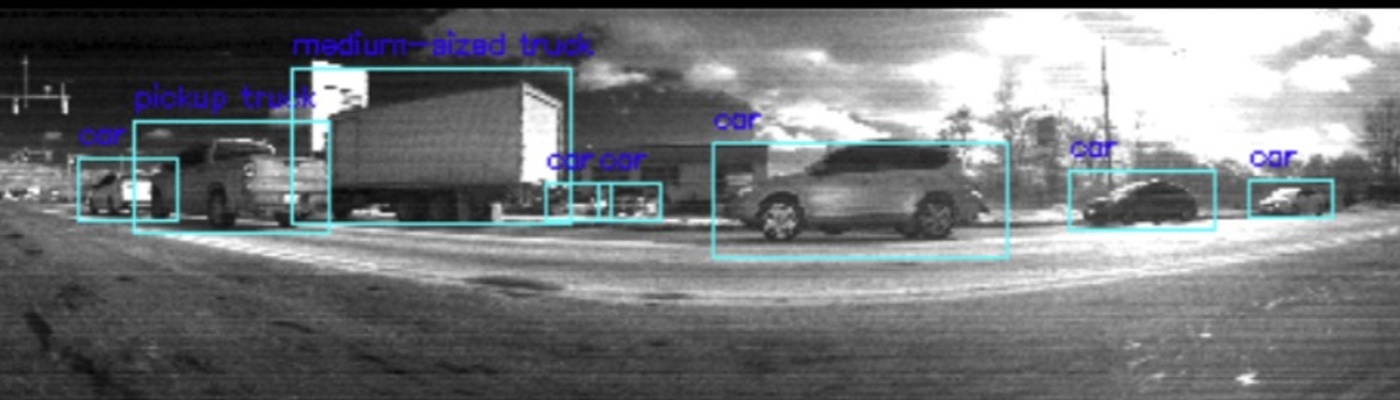}
    \\[1mm]
    (a) Both clear (NIR)
\end{minipage}
\hfill
\begin{minipage}[t]{0.32\textwidth}
    \centering
    \includegraphics[width=\textwidth]{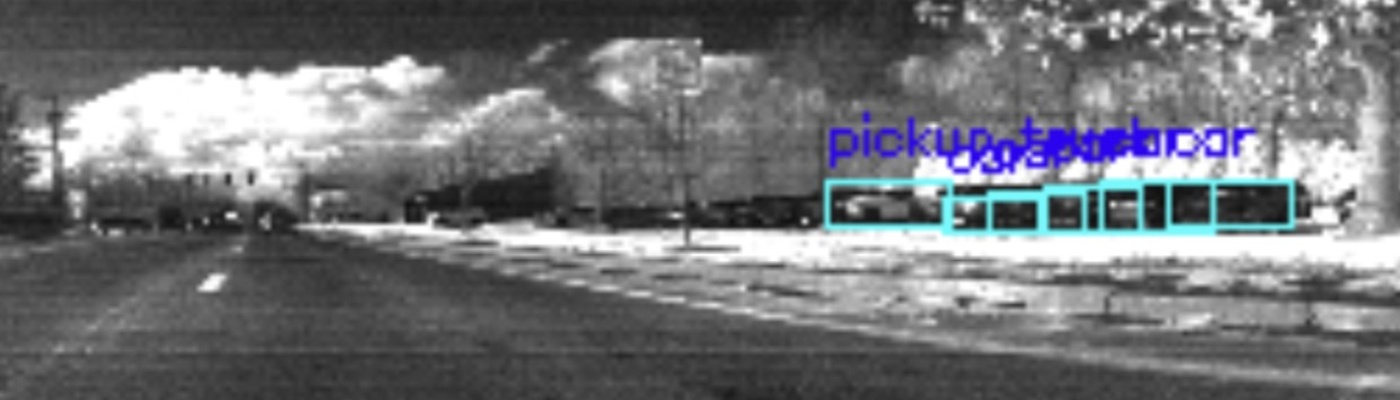}
    \\[1mm]
    (b) Point-dominant (NIR)
\end{minipage}
\hfill
\begin{minipage}[t]{0.32\textwidth}
    \centering
    \includegraphics[width=\textwidth]{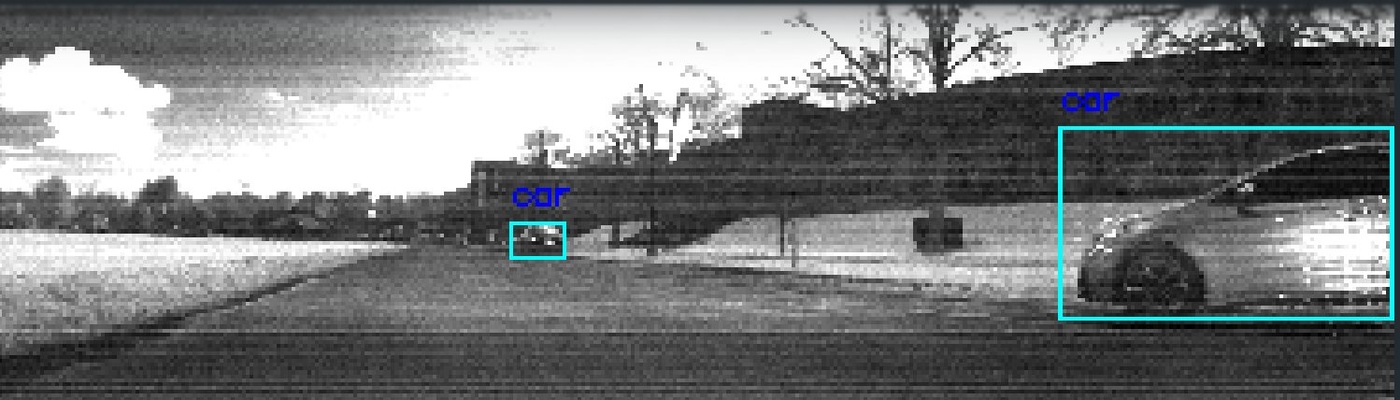}
    \\[1mm]
    (c) Image-dominant (NIR)
\end{minipage}

\vspace{2mm}

\begin{minipage}[t]{0.32\textwidth}
    \centering
    \includegraphics[width=\textwidth]{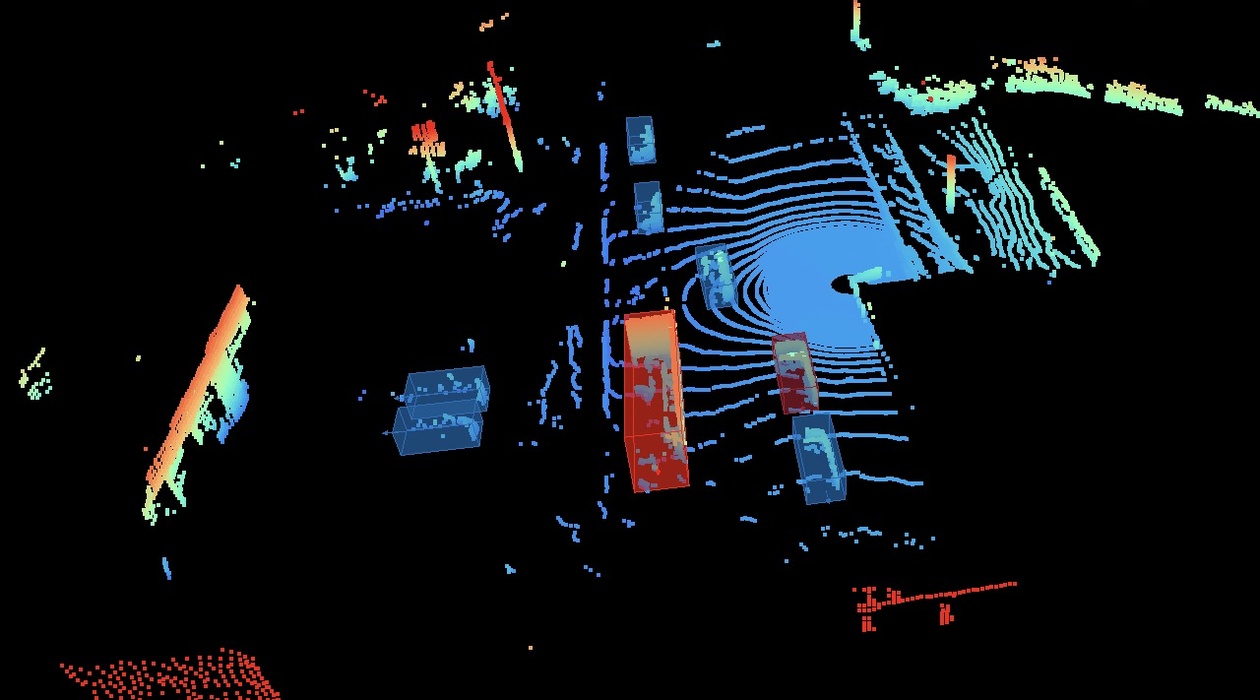}
    \\[1mm]
    (d) Both clear (Point Cloud)
\end{minipage}
\hfill
\begin{minipage}[t]{0.32\textwidth}
    \centering
    \includegraphics[width=\textwidth]{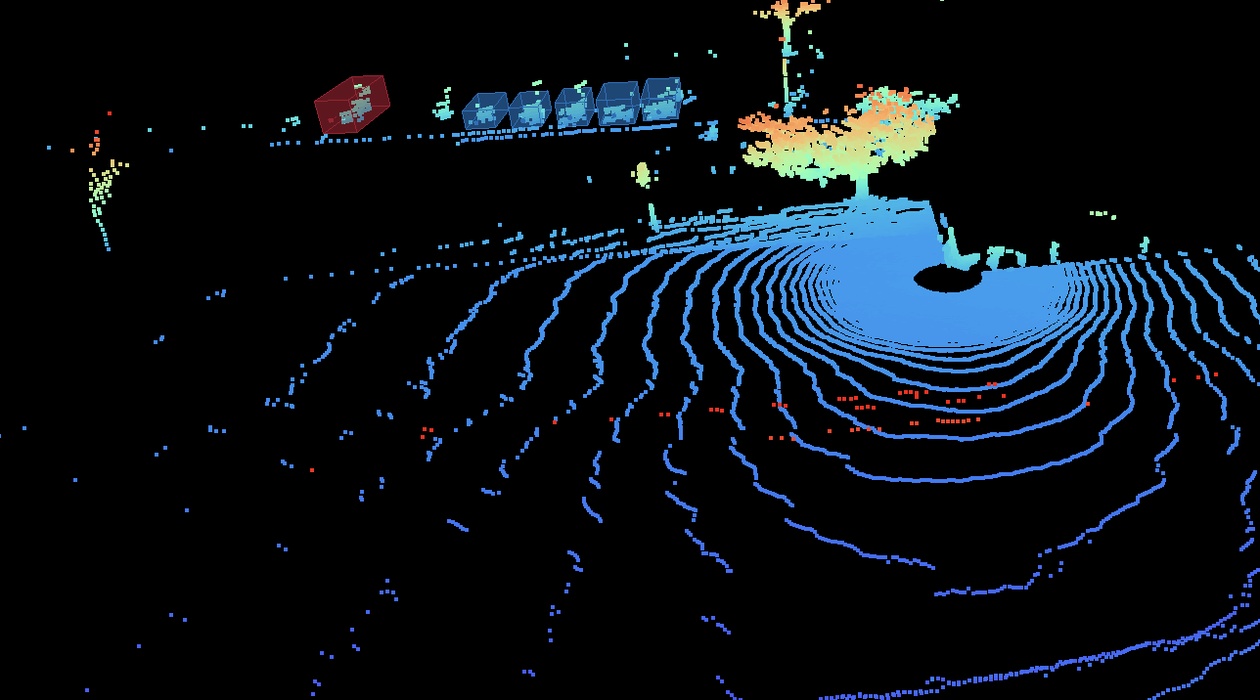}
    \\[1mm]
    (e) Point-dominant (Point Cloud)
\end{minipage}
\hfill
\begin{minipage}[t]{0.32\textwidth}
    \centering
    \includegraphics[width=\textwidth]{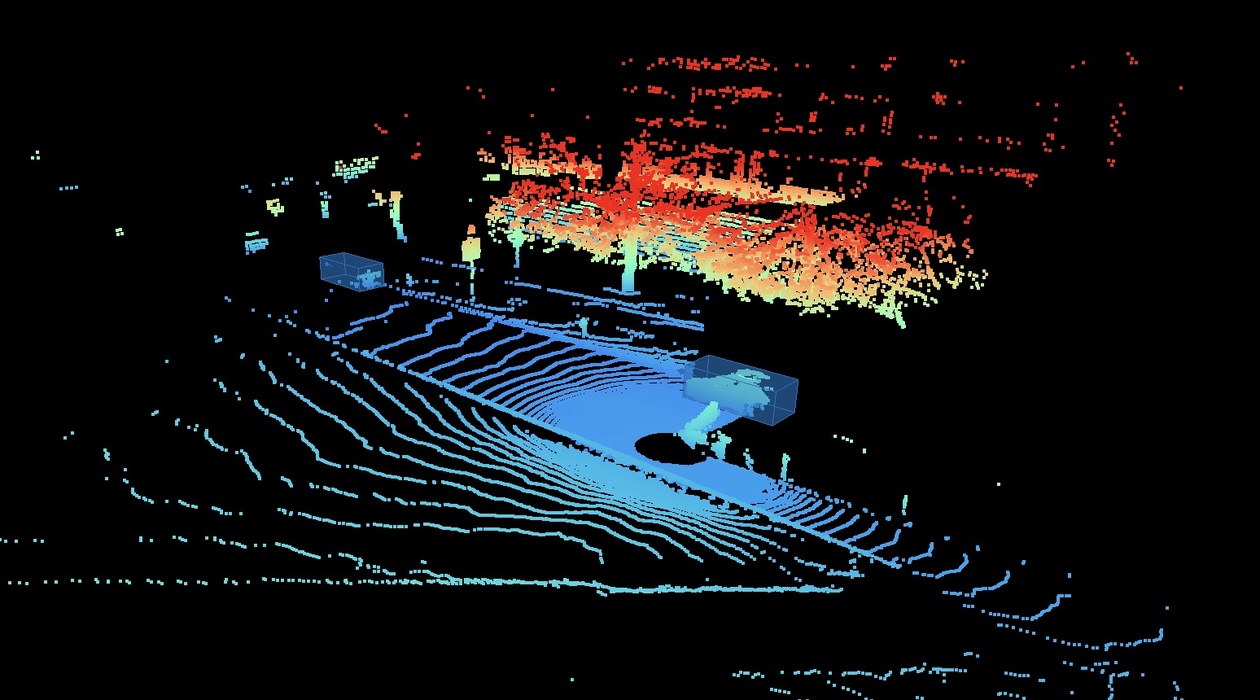}
    \\[1mm]
    (f) Image-dominant (Point Cloud)
\end{minipage}

\captionof{figure}{
Examples illustrating complementary characteristics between LiDAR geometric
and NIR image modalities. From left to right: (1) both modalities provide clear and consistent observations, (2) LiDAR point clouds dominate when vehicles appear overlapping in the 2D image plane due to occlusion, where 3D geometry enables unambiguous separation, and (3) NIR images dominate when point clouds are sparse or incomplete — due to limited returns at long range or partial scans of object surfaces — where image appearance cues enable reliable object recognition.}
\label{fig:app_modality_complementarity}

\end{center}

% ---- A.3 Category Ontology ----
\subsection{Category Ontology and Illustrations}
\label{app:ontology}

The PALIN ontology defines 11 object categories. Category boundaries are designed to maximize labeling consistency among visually or functionally similar vehicle types.

\textbf{Car.} Standard, light-duty passenger vehicles designed primarily for personal transport, including sedans, hatchbacks, station wagons, SUVs, crossovers, minivans, and passenger vans. This is the most frequent category in the dataset.

\textbf{Bus.} Full-size transit or coach buses with a large enclosed passenger compartment spanning the full vehicle length.

\textbf{Medium-size Truck.} Commercial utility vehicles with a box-shaped enclosed cargo body, including box trucks, step vans, and delivery trucks. Distinguished from pickup trucks by the absence of a separate open cargo bed.

\textbf{Pickup Truck.} Light-duty trucks with an enclosed passenger cabin and a separate open cargo bed. Distinguished from medium-size trucks by the exposed bed structure.

\textbf{Semi-Truck.} Articulated heavy-duty trucks consisting of a tractor unit and a detached trailer. Labeled as a single instance when the tractor and trailer are visually connected.

\textbf{Person.} Pedestrians, including walkers, joggers, and individuals standing near roadways. Excludes persons seated on bicycles or scooters (labeled separately).

\textbf{Cyclist.} Persons actively riding a bicycle, including both conventional and electric bicycles.

\textbf{Scooter.} Persons riding a motorized or electric scooter, including kick scooters.

\textbf{Motorcycle.} A motorized two-wheeled vehicle designed for road use.

\textbf{Train.} Large, rail-bound vehicles operating on fixed tracks, typically representing light rail, trams, or commuter trains.

\textbf{Towed Object.} Any unpowered vehicle, cargo trailer, or mobile equipment designed to be pulled by a motorized vehicle. This includes both actively towed objects and stationary towable items, such as boats parked in residential yards.  
 
Representative NIR image crops and point cloud views for each annotated object category in the PALIN dataset are available online.% \footnote{\href{https://github.com/moonmoonmoonmoon/PALIN/blob/main/Labelled_Dataset_Information.pdf}{Category illustrations}}

% ================= Fig. A3: Dataset Statistics =================
\begin{center}

\begin{minipage}[c]{0.48\linewidth}
    \centering
    \includegraphics[width=\linewidth]{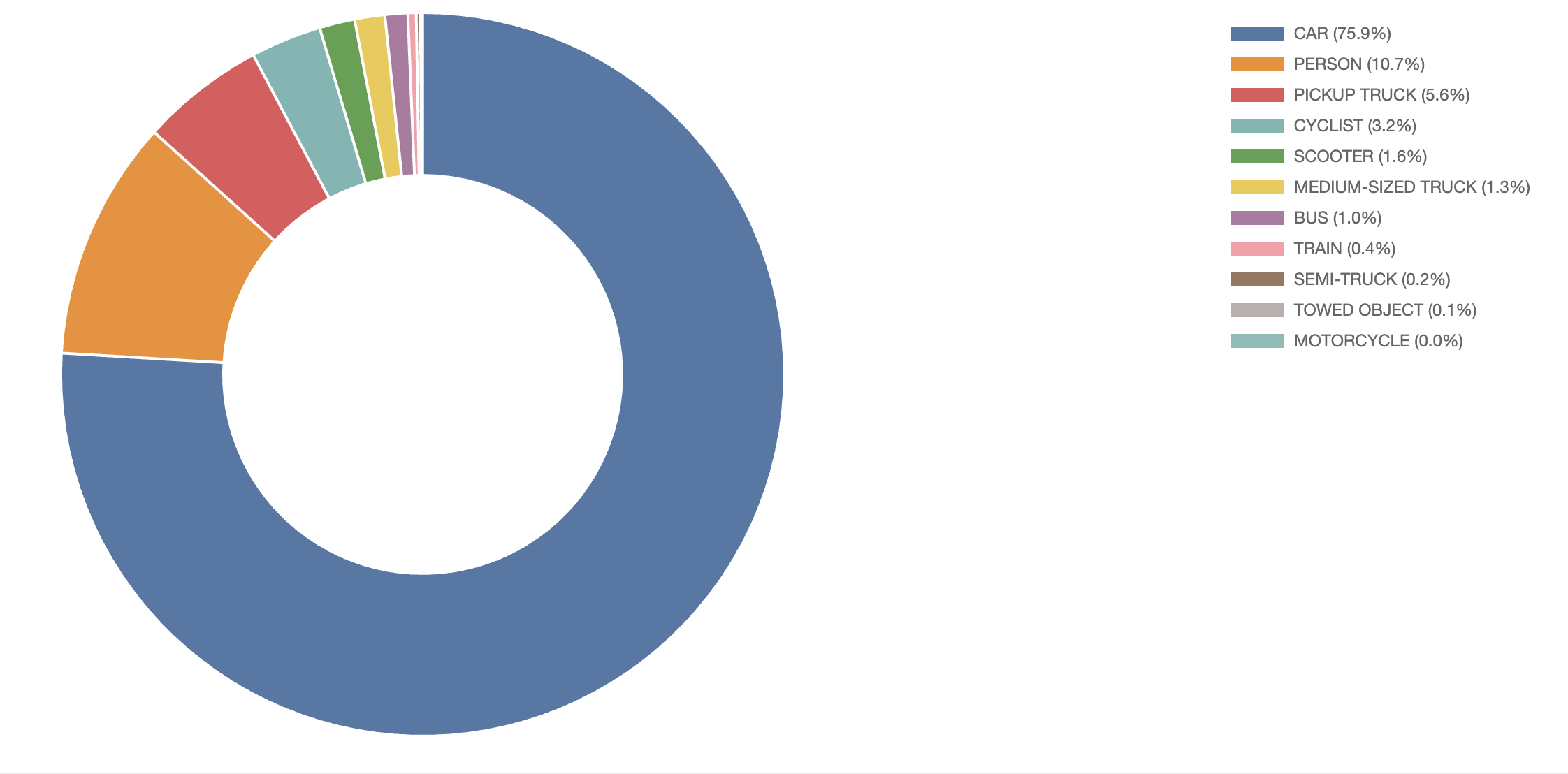}
    \\ (a) Category distribution.
\end{minipage}
\hfill
\begin{minipage}[c]{0.48\linewidth}
    \centering
    \includegraphics[width=\linewidth]{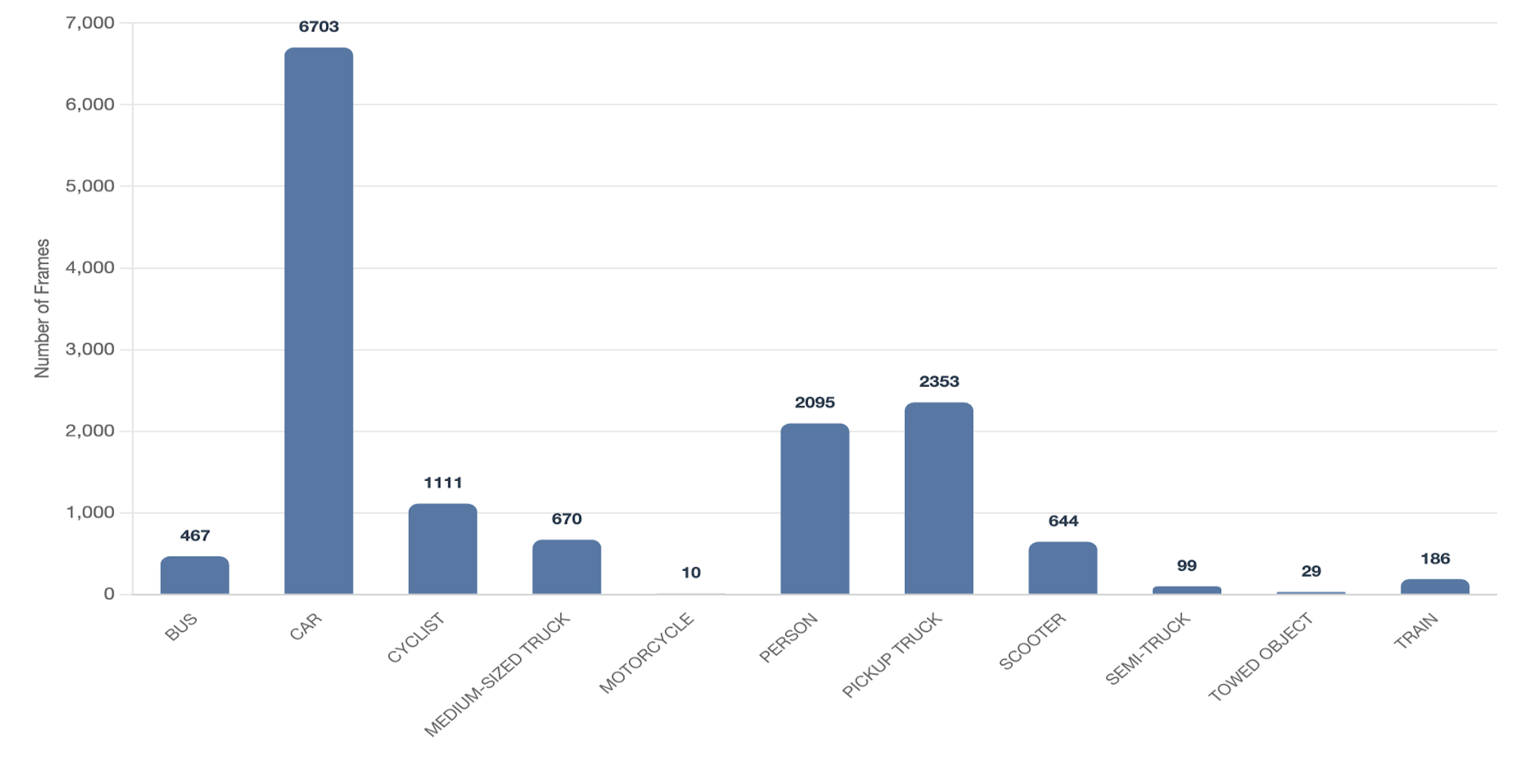}
    \\ (b) Frame count per class.
\end{minipage}

\vspace{2mm}

\begin{minipage}[c]{0.6\linewidth}
    \centering
    \includegraphics[width=\linewidth]{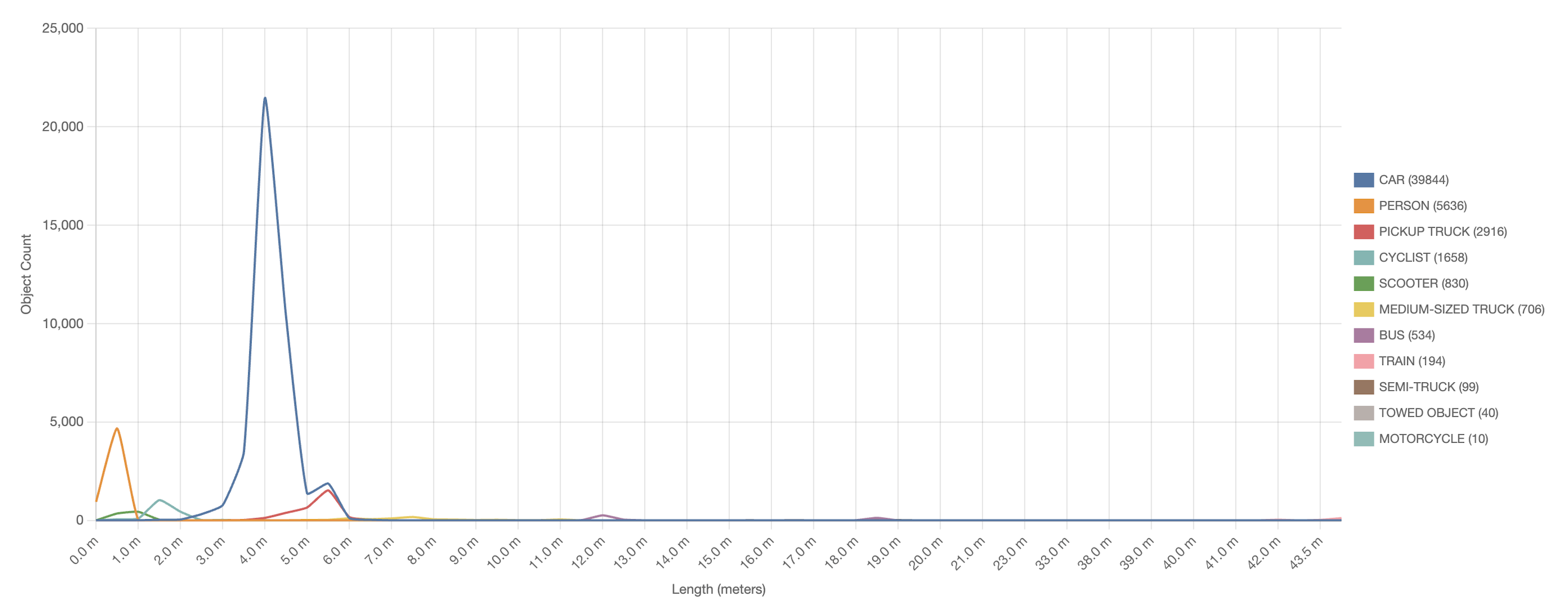}
    \\ (c) Object length distribution.
\end{minipage}

\captionof{figure}{
Dataset statistics of PALIN, including category distribution, per-class frame
count, and object length distribution.
}
\label{fig:app_dataset_statistics}

\end{center}

% ---- A.4 Category Ontology ----
\subsection{Comparison with public 3D datasets }
\label{app:app_dataset_comparison}
% ================= Table A1: Comparison of representative public 3D perception datasets =================
\begin{table}[H]
\centering
\caption{Comparison of representative public 3D perception datasets.}
\label{tab:app_dataset_comparison}

\setlength{\tabcolsep}{3pt}        
\renewcommand{\arraystretch}{1.08} 

\begin{threeparttable}
\small  

\begin{tabular}{lccllcccccl}
\toprule
Dataset & Year & Platform & \makecell{LiDAR\\Sensor} & \makecell{Image\\Modality}
  & \makecell{Native\\Align.}
  & $360^\circ$
  & \makecell{Ann.\\Frames\tnote{$\dagger$}}
  & Seqs.
  & \makecell{Det.\\Classes}
  & \makecell{Object Attributes} \\
\midrule

KITTI~\cite{kitti_cvpr,kitti}
  & 2012 & Vehicle
  & $1\times$ HDL-64E
  & RGB$\times$2 
  & $\times$ & $\times$
  & 7,481 & --- & 8
  & Occ., Trunc., Diff. \\

nuScenes~\cite{nuscenes}
  & 2020 & Vehicle
  & $1\times$ HDL-32E
  & RGB$\times$6 
  & $\times$ & $\checkmark$
  & 40,000 & 1,000 & 23
  & \makecell[l]{Occ., Motion} \\

Waymo~\cite{waymo}
  & 2020 & Vehicle
  & $5\times$ LiDAR (fused)
  & RGB$\times$5
  & $\times$ & $\checkmark$
  & 200,000 & 1,000 & 4
  & Diff.(L1/L2), Heading \\

PandaSet~\cite{pandaset}
  & 2021 & Vehicle
  & $2\times$ Hesai Pandar64 
  & RGB$\times$6
  & $\times$ & $\checkmark$
  & 8,240 & 103 & 28
  & Occ., Heading \\

\midrule

\textbf{PALIN (Ours)}
  & \textbf{2026} & \textbf{Bus + Road}
  & $\mathbf{2\times}$ \textbf{OS1-128}\tnote{$\ddagger$}
  & \textbf{NIR}
  & $\checkmark$ & $\checkmark$
  & \textbf{8,587} & $\mathbf{6}$\tnote{$\S$} & \textbf{11}
  & \makecell[l]{\textbf{Occ., Diff.\ (E/M/H), Yaw}\\
      \textbf{+ \underline{Modality Dominance}}\tnote{$\P$}} \\

\bottomrule
\end{tabular}

\begin{tablenotes}
\footnotesize
\item[$\dagger$] Annotated LiDAR 3D detection frames with public ground truth only. KITTI test split (7,518 frames) has no public labels. nuScenes annotates keyframes at 2\,Hz from 20\,Hz raw streams; Waymo and PandaSet annotate all captured frames.
\item[$\ddagger$] 3D annotations are assigned to the left-mounted sensor's coordinate frame; both sensors are released.
\item[$\S$] Sequences are continuous recording sessions (bus routes or fixed deployments), not manually curated short clips as in nuScenes or Waymo.
\item[$\P$] Modality dominance attributes indicate whether each annotated object is primarily observable from LiDAR geometry, NIR appearance, or both, enabling fine-grained evaluation of modality complementarity.
\end{tablenotes}

\end{threeparttable}
\end{table}
\clearpage

% ================= B =================
\section{CMambaFusion Algorithm}
\label{app:algorithm}

\renewcommand{\thealgorithm}{B\arabic{algorithm}}

\setcounter{algorithm}{0}
% ================= Algorithm B1 =================

% \FloatBarrier
\begin{algorithm}[H]
\centering
\noindent\rule{\textwidth}{0.4pt}
\vspace{-1mm}

\textbf{4-Dir Cross-Mamba \hspace{3.7cm} Gated Concat-Mamba}

\vspace{1mm}
\noindent\rule{\textwidth}{0.4pt}
\vspace{2mm}

\begin{minipage}[t]{0.475\textwidth}
\begin{algorithmic}[1]
\Require $F_g,\, F_n \in \mathbb{R}^{H_f \times W_f \times d}$
\Ensure $\tilde{F}_g,\, \tilde{F}_n \in \mathbb{R}^{H_f \times W_f \times d}$
\State $X_g \gets \mathrm{DWConv}_g\!\left(\phi_g(F_g)\right)$
\State $X_n \gets \mathrm{DWConv}_n\!\left(\phi_n(F_n)\right)$
\State $\{u_g^k\}_{k=1}^{4} \gets \mathrm{Scan4D}(X_g)$
\State $\{u_n^k\}_{k=1}^{4} \gets \mathrm{Scan4D}(X_n)$
\For{$k = 1, \ldots, 4$}
    \State $y_g^k \gets \mathrm{SSM}(u_g^k;\; C_n^k)$ \Comment{$C$-matrix swap}
    \State $y_n^k \gets \mathrm{SSM}(u_n^k;\; C_g^k)$
\EndFor
\State $\tilde{F}_g \gets F_g + \psi_g\!\left(\mathrm{LN}(\mathrm{merge}(\{y_g^k\}))\right)$
\State $\tilde{F}_n \gets F_n + \psi_n\!\left(\mathrm{LN}(\mathrm{merge}(\{y_n^k\}))\right)$
\end{algorithmic}
\end{minipage}
\hfill
\vrule
\hfill
\begin{minipage}[t]{0.475\textwidth}
\begin{algorithmic}[1]
\Require $\tilde{F}_g,\, \tilde{F}_n \in \mathbb{R}^{H_f \times W_f \times d}$
\Ensure $F_{\mathrm{out}} \in \mathbb{R}^{H_f \times W_f \times d}$
\State $X_0 \gets \mathrm{ReLU}\!\left(\mathrm{BN}\!\left(\mathrm{Conv}_{1\times1}([\tilde{F}_g;\tilde{F}_n])\right)\right)$ \Comment{dim $d$}
\State $Z \gets \mathrm{ReLU}\!\left(\mathrm{BN}\!\left(\mathrm{Conv}_{1\times1}([\tilde{F}_g;\tilde{F}_n])\right)\right)$ \Comment{dim $rd$}
\State $\{v^k\}_{k=1}^{4} \gets \mathrm{Scan4D}(\mathrm{DWConv}(Z))$
\For{$k = 1, \ldots, 4$}
    \State $w^k \gets \mathrm{SSM}(v^k)$ \Comment{single-stream}
\EndFor
\State $Y \gets \mathrm{merge}(\{w^k\}_{k=1}^{4})$
\State $g \gets \sigma\!\left(\mathrm{Conv}_{1\times1}(\tilde{F}_g - \tilde{F}_n)\right)$
\State $F_{\mathrm{out}} \gets X_0 + g \odot Y$
\end{algorithmic}
\end{minipage}

\vspace{2mm}
\noindent\rule{\textwidth}{0.4pt}
\caption{\textbf{: CMambaFusion.}  \textit{Left}: 4-Dir Cross-Mamba. \textit{Right}: Gated Concat-Mamba. \\ $\phi$: linear projection; $\psi$: output projection; 
LN: layer norm; $r$: SSM expansion ratio.}
\label{alg:cmambafusion}
\end{algorithm}

% ================= C =================
\section{Evaluation Protocol}
\label{app:impl}
 
\renewcommand{\thefigure}{C\arabic{figure}}
\renewcommand{\thetable}{C\arabic{table}}
\setcounter{figure}{0}
\setcounter{table}{0}
 
% \subsection{Evaluation Protocol}
We adopt 3D Average Precision with 40 recall positions (AP$_{40}$)~\cite{kitti,simonelli}, which provides a more stable and fine-grained approximation of the precision--recall curve than the earlier 11-point sampling and has become the standard in 3D object detection.
 
IoU thresholds follow the KITTI convention~\cite{kitti}: 0.7 and 0.5 are applied to vehicle-type categories, and 0.5 and 0.25 to small object categories (Pedestrian, Cyclist, and Scooter). For PALIN-specific categories (Train, Semi-Truck, Pickup Truck, Medium Truck, Bus), vehicle-type thresholds are applied based on object scale. Per-class AP$_{40}$ and weighted mAP$_{40}$ are reported at both IoU thresholds.
 
Results are reported for 8 evaluable categories with sufficient training samples; categories with extremely limited instances (e.g., Motorcycle, Towed Object) are excluded to ensure statistically reliable evaluation. The best checkpoint is selected based on weighted mAP$_{40}$ over the final training epochs, where the class weights are proportional to per-class instance counts normalized to sum to one.

% ================= D =================
\section{Additional Results and Visualization}
\label{app:qualitative}

\renewcommand{\thefigure}{D\arabic{figure}}
\renewcommand{\thetable}{D\arabic{table}}
\setcounter{figure}{0}
\setcounter{table}{0}

% ================= Table D1 =================

\begin{center}

\captionof{table}{Per-class 3D AP$_{40}$ (\%) on the PALIN Full dataset (8k).}
\label{tab:perclass}

\setlength{\tabcolsep}{4.5pt}
\renewcommand{\arraystretch}{1.15}

\resizebox{\textwidth}{!}{%
\begin{threeparttable}
\begin{tabular}{lcccccccc}
\toprule
\multirow{2}{*}{Class}
& \multicolumn{2}{c}{PointPillars}
& \multicolumn{2}{c}{Cross-Attn}
& \multicolumn{2}{c}{4-Dir Cross-Mamba}
& \multicolumn{2}{c}{CMambaFusion} \\
\cmidrule(lr){2-3}\cmidrule(lr){4-5}\cmidrule(lr){6-7}\cmidrule(lr){8-9}
& @0.7 & @0.5 & @0.7 & @0.5 & @0.7 & @0.5 & @0.7 & @0.5 \\
\midrule

Car
& 74.24 & 90.85
& 76.57 & 92.60
& 79.05 & 93.02
& \textbf{79.49} & \textbf{93.24} \\

Pedestrian$^\dagger$
& 55.66 & 66.82
& 51.20 & 70.95
& 56.49 & 72.72
& \textbf{56.83} & \textbf{74.98} \\

Bus
& 74.04 & 86.89
& 70.62 & 81.92
& \textbf{74.27} & 84.50
& 71.48 & \textbf{85.00} \\

Pickup Truck
& 67.18 & 77.19
& 59.37 & 70.07
& 54.97 & 69.99
& \textbf{62.47} & \textbf{73.76} \\

Medium Truck
& 83.31 & 87.36
& 55.70 & 80.31
& 53.91 & 76.99
& \textbf{61.01} & \textbf{82.71} \\

Cyclist$^\dagger$
& 72.31 & 80.11
& 69.24 & 78.33
& \textbf{71.72} & \textbf{82.32}
& 70.15 & 79.16 \\

Scooter$^\dagger$
& 49.00 & 65.60
& 50.33 & 54.86
& 56.08 & 60.99
& \textbf{56.54} & \textbf{64.81} \\

Train
& 29.10 & 55.00
& \textbf{52.39} & 57.31
& 42.39 & \textbf{57.50}
& 52.17 & 55.00 \\

\bottomrule
\end{tabular}

\begin{tablenotes}
  \footnotesize
  \item \textbf{Bold} highlights the best performance among fusion methods (excluding the PointPillars baseline).
  \item[$\dagger$] Small-object categories evaluated at IoU thresholds of 0.5/0.25, reported in the @0.7/@0.5 columns for consistency.
\end{tablenotes}

\end{threeparttable}
}

\end{center}

% ================= Fig. D1: Qualitative Examples =================
\begin{center}

\begin{minipage}[c]{\linewidth}
    \centering
    \includegraphics[width=\linewidth]{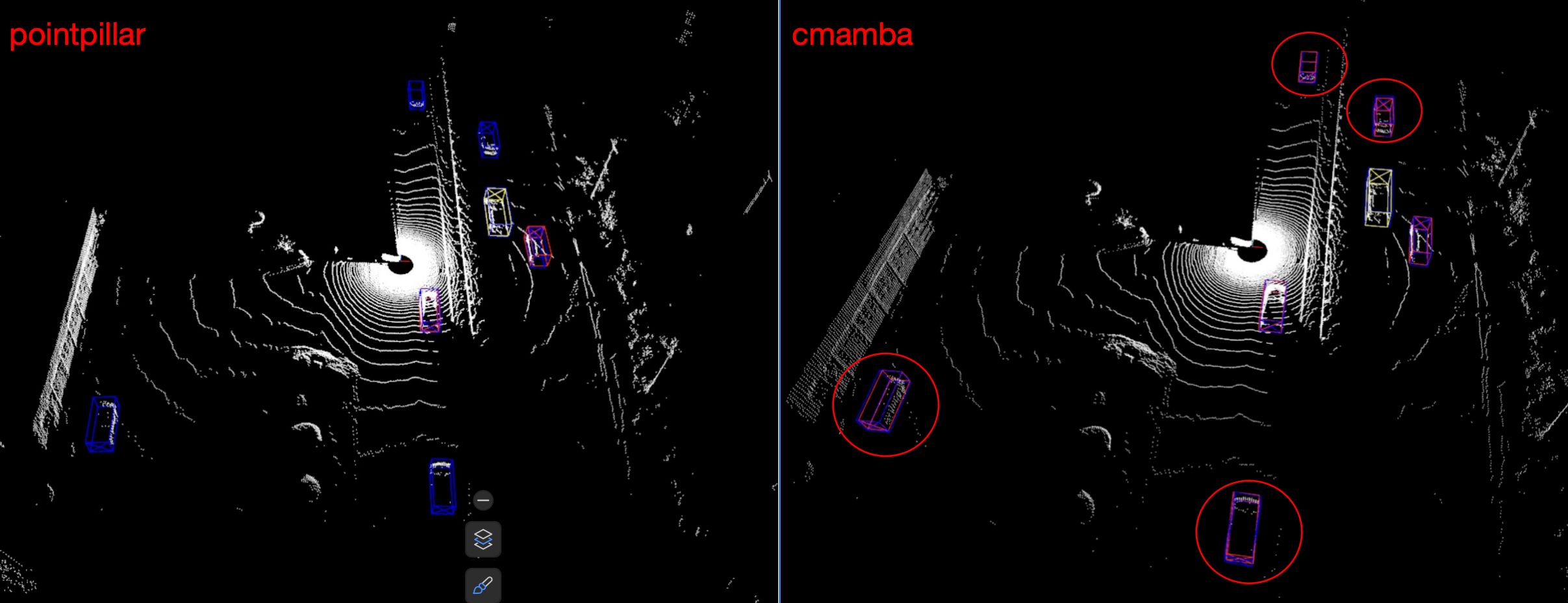}
    \\ (a) Bus-mounted dynamic scene.
\end{minipage}

\vspace{2mm}

\begin{minipage}[c]{\linewidth}
    \centering
    \includegraphics[width=\linewidth]{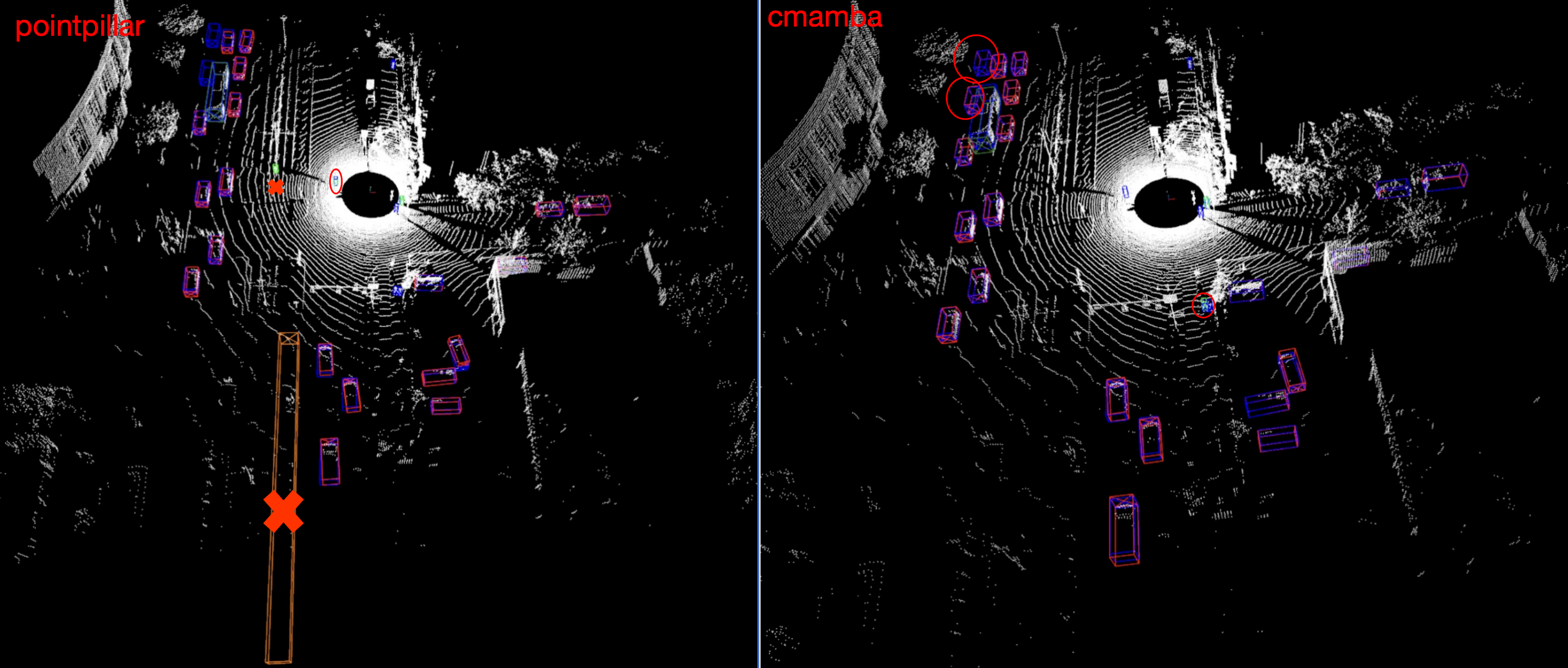}
    \\ (b) Boston static intersection scene.
\end{minipage}

\captionof{figure}{
Qualitative comparison between the LiDAR-only baseline and CMambaFusion.
Blue boxes denote ground-truth annotations, and colored boxes represent
predicted classes. Red circles highlight additional true positives (+TP) detected by CMambaFusion but missed by PointPillars, while red crosses (×) indicate false positives (FP).}
\label{fig:app_qualitative}

\end{center}

\end{document}